\documentclass[10pt,twocolumn,letterpaper]{article}

\usepackage{wacv}
\usepackage{times}
\usepackage{epsfig}
\usepackage{graphicx}
\usepackage{amsmath}
\usepackage{amssymb}

\usepackage{caption}
\usepackage{subcaption}
\usepackage{array,multirow}
\usepackage{threeparttable}
\usepackage{booktabs}
\usepackage{makecell}
\usepackage{tabularx}
\usepackage[accsupp]{axessibility}
\usepackage[toc,page]{appendix}

\usepackage[pagebackref=true,breaklinks=true,letterpaper=true,colorlinks,bookmarks=false]{hyperref}

\def\wacvPaperID{0934} 

\wacvfinalcopy 

\ifwacvfinal
\fi

\ifwacvfinal
\usepackage[breaklinks=true,bookmarks=false]{hyperref}
\else
\usepackage[pagebackref=true,breaklinks=true,colorlinks,bookmarks=false]{hyperref}
\fi

\ifwacvfinal
\fi

\usepackage[dvipsnames]{xcolor}
\definecolor{s1}{RGB}{157, 179, 156}
\definecolor{s2}{RGB}{205, 176, 147}
\definecolor{s3}{RGB}{163, 179, 202}

\begin{document}

\title{Human-Aided Saliency Maps Improve Generalization of Deep Learning}

\author{Aidan~Boyd,~Kevin~Bowyer,~Adam~Czajka\\
University of Notre Dame, 
Notre Dame, IN 46556\\
{\tt\small \{aboyd3,kwb,aczajka\}@nd.edu}
}

\maketitle

\ifwacvfinal
\thispagestyle{empty}
\fi

\begin{abstract}
Deep learning has driven remarkable accuracy increases in many computer vision problems. One ongoing challenge is how to achieve the greatest accuracy in cases where  training data is limited. A second ongoing challenge is that trained models oftentimes do not generalize well even to new data that is subjectively similar to the training set. We address these challenges in a novel way, with the first-ever (to our knowledge) exploration of encoding human judgement about salient regions of images into the training data. We compare the accuracy and generalization of a state-of-the-art deep learning algorithm for a difficult problem in biometric presentation attack detection when trained on (a) original images with typical data augmentations, and (b) the same original images transformed to encode human judgement about salient image regions. 
The latter approach results in models that achieve higher accuracy and better generalization, decreasing the error of the LivDet-Iris 2020 winner from 29.78\% to 16.37\%, and achieving impressive generalization in a leave-one-attack-type-out evaluation scenario. This work opens a new area of study for how to embed human intelligence into training strategies for deep learning to achieve high accuracy and generalization in cases of limited training data.

\end{abstract}

\begin{figure}[t]
    \centering
    \includegraphics[width=1\columnwidth]{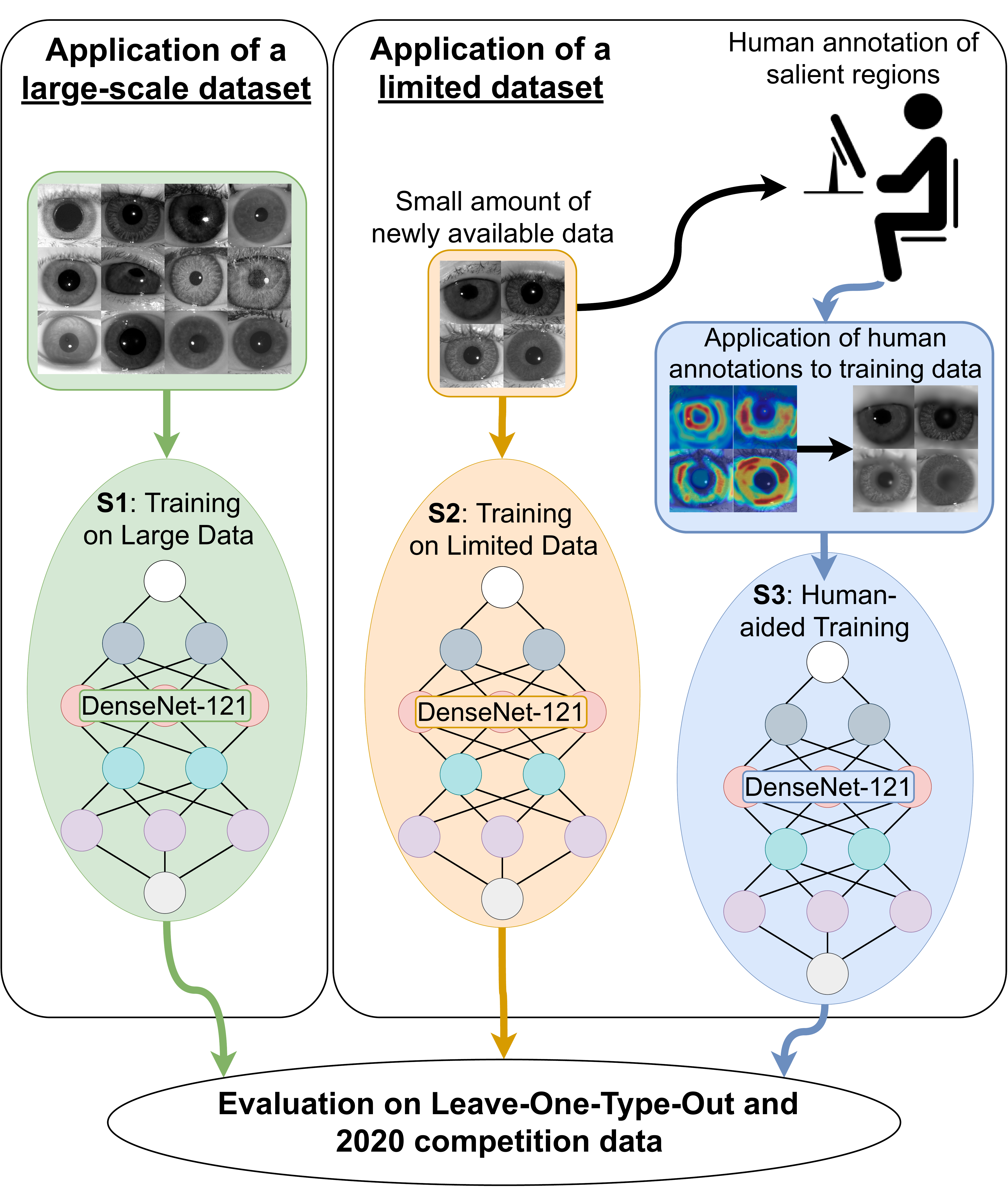}
    \caption{{\bf Do Human-Aided Saliency Maps Improve Generalization of Deep Learning?} 
We use the same training and architecture to compare models trained on a newly-acquired limited dataset, with typical augmentation techniques (\textcolor{s2}{\bf S2}), and the same images encoding human judgement of region saliency using multiple levels of blur (\textcolor{s3}{\bf S3}).
For reference, we also present results for training with a much larger original dataset (\textcolor{s1}{\bf S1}).
The use of training images encoding human saliency results in models demonstrating {\bf improved accuracy and generalization over the conventional approach}.} \vskip-6mm
    \label{teaser}
\end{figure}

\section{Introduction}

Deep learning methods have had huge impact in many areas of computer vision,
including generic object detection \cite{Liu_IJCV_2020}, 
face recognition \cite{Masi_2018, Guo_2019},
medical image analysis \cite{Litjens_2017} and
biometrics \cite{Sundararajan_CSUR_2018}.
They are known for their need for large amounts of training data and for solutions that are often fragile, in the sense of producing state-of-the-art results on a well-defined problem but 
not generalizing well to what might seem to be related problems or datasets.
The generalization problem can be addressed to some degree by design of a large and varied training dataset, or regularization techniques.
But the strengths and weaknesses here are like two sides of the deep learning coin.
Deep learning-based methods can learn whatever exists in the training data that can be used to solve the problem. Learning what is
incidental to the training data causes the solution to not generalize well.

In this work, we make the novel proposition to transform training data for deep learning in a way that incorporates {\bf human judgement} about salient parts of images. We asked humans to annotate image regions that are salient to their decision about that image. Then we produced transformed versions of the original images, in which the degree of blur applied to a region is inversely related to its human-rated salience. The transformed images have full original clarity in regions salient to humans, and increasing levels of blur in regions rated increasingly less salient for humans.
Such transformed images are then used as training data. In this way, the deep learning process is encouraged to learn whatever it can from the image regions rated salient for humans, but discouraged from learning from high-frequency details of regions rated less salient for humans, 
Fig. \ref{teaser}. 

Various methods of experimental psychology have been already used to gauge limits of human perception \cite{Li_JV_2007,RichardWebster_TPAMI_2018}, understand how humans process visual information \cite{Berga_ELCVIA_2020}, build more explainable computer vision models \cite{RichardWebster_ECCV_2018}, or strengthen feature extractors in iris recognition with human-driven way of processing images \cite{Czajka_WACV_2019}. However, to our knowledge, this is the first-ever exploration of transforming the training data for deep learning based on human perception.
This may initially seem a radical proposition. But we present two important advantages of this approach when applied to a quite difficult problem of presentation attack detection in biometrics, which always suffers from small and unrepresentative training corpora. First, comparing the accuracy achieved by training with the original image dataset versus the saliency-transformed version of that dataset, the saliency-transformed version achieves significantly higher accuracy. Second, evaluating the open-set scenario of training using all-except-one attack types and testing on the held-out attack type, the saliency-transformed training generates models with significantly higher accuracy on generalizing to the held-out attack type. It's noteworthy that the idea proposed in this paper is different from and independent of recent research on ``attention'' mechanisms \cite{Bahdanau_ICLR_2015,Bello_ICCV_2019}, and actually may complement them with human guidance.

This work opens up a new area of study for how to best incorporate human judgement about visual saliency into training data for deep learning, and makes the following {\bf novel contributions}:\vskip1mm
(a) a framework that {\bf employs human intelligence to use effectively a low amount of training data} to increase the performance of deep learning models, \vskip1mm
(b) demonstrated {\bf greater generalization} of the method when applied to the iris presentation attack detection problem being solved by a deep learning classifier,\vskip1mm
(c) all elements to reproduce this work made publicly available, including a {\bf database of human-annotated iris images and sources codes}.

\section{Related Work}
\begin{table*}[!h]
    \begin{center}
    \caption{Number of samples in the {\bf train}, {\bf validation} and {\bf test} partitions extracted from the superset of all available iris PAD datasets, broken by abnormality type. [P] denotes our proprietary data set.}
    \label{tab:dataset}
    \footnotesize
    \begin{tabular}{|c|c|c|c|c|c|c|c|c|c|}
        \hline
        \textbf{Bona fide} & \textbf{Artificial} & \textbf{Textured} & \textbf{Display} & \textbf{Post} & \textbf{Paper} & \textbf{Synthetic} & \textbf{Diseased} & \textbf{Textured contact} & \textbf{Total}\\
        & & \textbf{contact lenses} & & \textbf{mortem} & \textbf{printouts} & & & \textbf{lenses \& printed} & \\\hline\hline
        \multicolumn{10}{|c|}{Train and validation partitions:}\\
        399,053 & 277 & 27,372 & $\times$ & 2,259 & 16,393 & 10,000 & 1,537 & 1,899 & {\bf 458,790} \\
        \cite{casia-database} \cite{Sung_OE_2007} \cite{Galbally_ICB_2012} & \cite{Sung_OE_2007} [P] &
        \cite{Sung_OE_2007} \cite{Kohli_ICB_2013} \cite{Yambay_ISBA_2017} & & \cite{Trokielewicz_IVC_2020} & 
        \cite{Galbally_ICB_2012} \cite{Sung_OE_2007} \cite{Kohli_BTAS_2016} & \cite{Wei_ICPR_2008} & \cite{Trokielewicz_BTAS_2015} & \cite{Yambay_IJCB_2017} & \\
        \cite{Trokielewicz_BTAS_2015} \cite{ETPAD_v2_URL} \cite{Kohli_ICB_2013} & & 
        \cite{Yambay_IJCB_2017} [P] & & & & & & & \\
        \cite{Yambay_ISBA_2017} \cite{WARSAW_DBs_URL} \cite{Yambay_IJCB_2017} [P] & & & & & & & & & \\\hline\hline
        \multicolumn{10}{|c|}{Test partition (equivalent to LivDet-Iris 2020 benchmark) \cite{Das_IJCB_2020}:}\\
        5,331 & 541 & 4,336 & 81 & 1,094 & 1,049 & $\times$ & $\times$ & $\times$ & \textbf{12,432} \\\hline
    \end{tabular}
    \end{center}
\end{table*}
\begin{figure*}[t]
    \centering
    \includegraphics[width=1\linewidth]{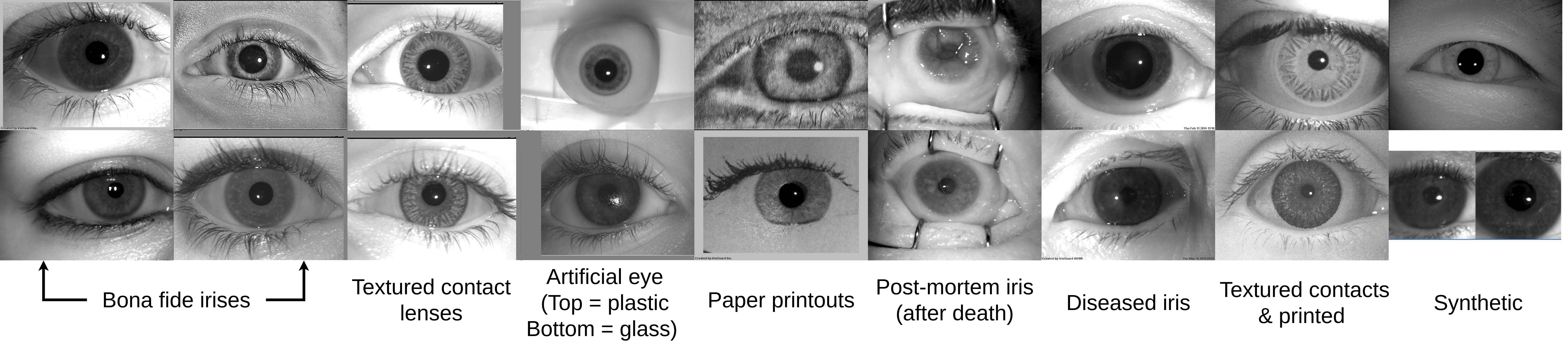}
    \caption{Examples of {\it bona fide} and {\it abnormal} samples from the acquired databases. The eight image types 
    shown represent all attacks represented in the datasets. Each human annotator was presented with multiple examples of each type.}
    \label{fig:attacks}
\end{figure*}

\noindent{\bf Training data augmentation.} 
Deep learning’s voracious appetite for training datasets has been dealt with in multiple ways,
and data augmentation techniques (such as flipping orientation of images, adding noise, using multiple crop offsets, and combination of those) are now a staple element of training \cite{Shorten_JBD_2019,Cubuk_CVPR_2019,Luo_CVPR_2020}.
These techniques can be applied to raw training data, subsets of training data within a batch \cite{Hoffer_CVPR_2020}, or in feature space instead of image space \cite{Ko_CVPR_2020}. Another approach, benefiting from a renaissance of generative models, is to 
generate large amounts of synthetic images for use in training \cite{Tsirikoglou_CGF_2020}. Recent trends are to mix synthetic and actual data \cite{Koutilya_CVPR_2020}, \eg pre-train a model on a larger synthetically-generated corpus, and fine-tune on a smaller actual sample set \cite{Wang_CVPR_2019}, or guide the generation of synthetic data to make it more domain-specific \cite{Kar_ICCV_2019}. 

To our knowledge, no previous work has 
applied human saliency encoding to augmentation techniques.

\vskip1mm\noindent{\bf Combining human and algorithm capabilities} when solving vision tasks has been studied in the context of biometrics. O'Toole \etal \cite{OToole_TSMC_2007} demonstrated that fusing humans and algorithms increased face recognition accuracy to near perfect values for the Face Recognition Grand Challenge dataset. A few research groups concluded that human's and algorithm's visual saliencies differ and their integration increases the quality of image captioning \cite{He_ICCV_2019} and of post-mortem iris recognition \cite{Moreira_WACV_2019,Trokielewicz_BTAS_2019}. 
Peterson \etal have even shown that humans' perceptual uncertainty may positively impact the generalization of deep learning-based models \cite{Peterson_ICCV_2019}. 
Human-machine pairing was also proposed to speed-up large-scale segmentation tasks, in which humans ``correct'' algorithm results, without a need to solve an entire segmentation task manually \cite{Benenson_CVPR_2019}. 

These past efforts demonstrate a boost in performance when machines and humans cooperate on the same task. However, again, we know of no previous work that has 
integrated human saliency judgements into training to increase generalization of the models.

\vskip1mm\noindent{\bf Presentation Attack Detection (PAD)} refers to determining the validity of an object presented to a biometric sensor, which -- if not detected -- 
could drive the system to an incorrect decision.
While many iris PAD works show promising cross-dataset and cross-attack performance, generalization against true {\it unknown} attack types is an open research problem and a crucial aspect of deployable solutions \cite{Boyd_PRL_2020}. Modern iris PAD approaches mainly employ deep-learning to achieve state-of-the-art accuracy, as evidenced in the recent LivDet-Iris 2020 competition \cite{Das_IJCB_2020}. 
In particular, Sharma and Ross \cite{Sharma_IJCB_2020} propose the application of DenseNet-121 \cite{Huang_2017_CVPR} to iris PAD with a focus on interpretability. 
More recently, Chen and Ross proposed a novel method of attention-guided training using class activation mapping and attention modules \cite{Chen_WACVW_2021}. They apply positional and channel attention modules to extract refined features from a DenseNet-121 backbone. Through experiments on public and private datasets, superior performance is demonstrated on both known and unknown presentation attack samples.
The attention modules shift the focus of the networks to the iris regions rather than peripheral image features. 
Our work differs 
in that, instead of applying attention modules to learn network-defined salient features, the collected human-annotated features are used as predefined salient features and incorporated directly into the training process. 

\section{Experimental Datasets}

\begin{figure*}[!ht]
    \centering
    \begin{subfigure}[b]{1\textwidth}
        \begin{subfigure}[b]{0.218\textwidth}
          \centering
          \includegraphics[width=1\columnwidth]{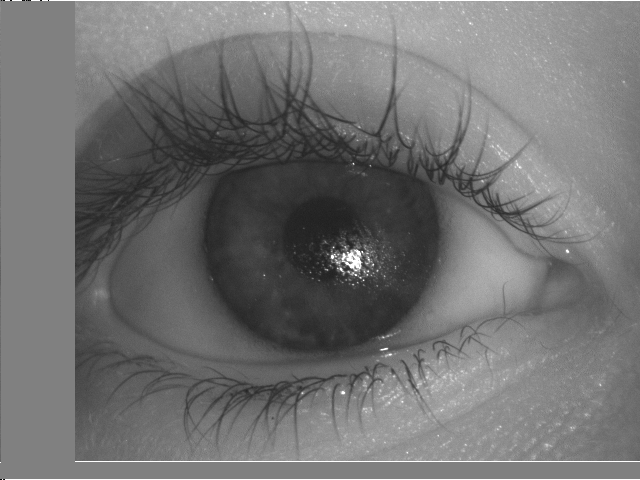}
          \caption{~}
        \end{subfigure}
        \hfill
        \begin{subfigure}[b]{0.326\textwidth}
            \centering
            \includegraphics[width=1\columnwidth]{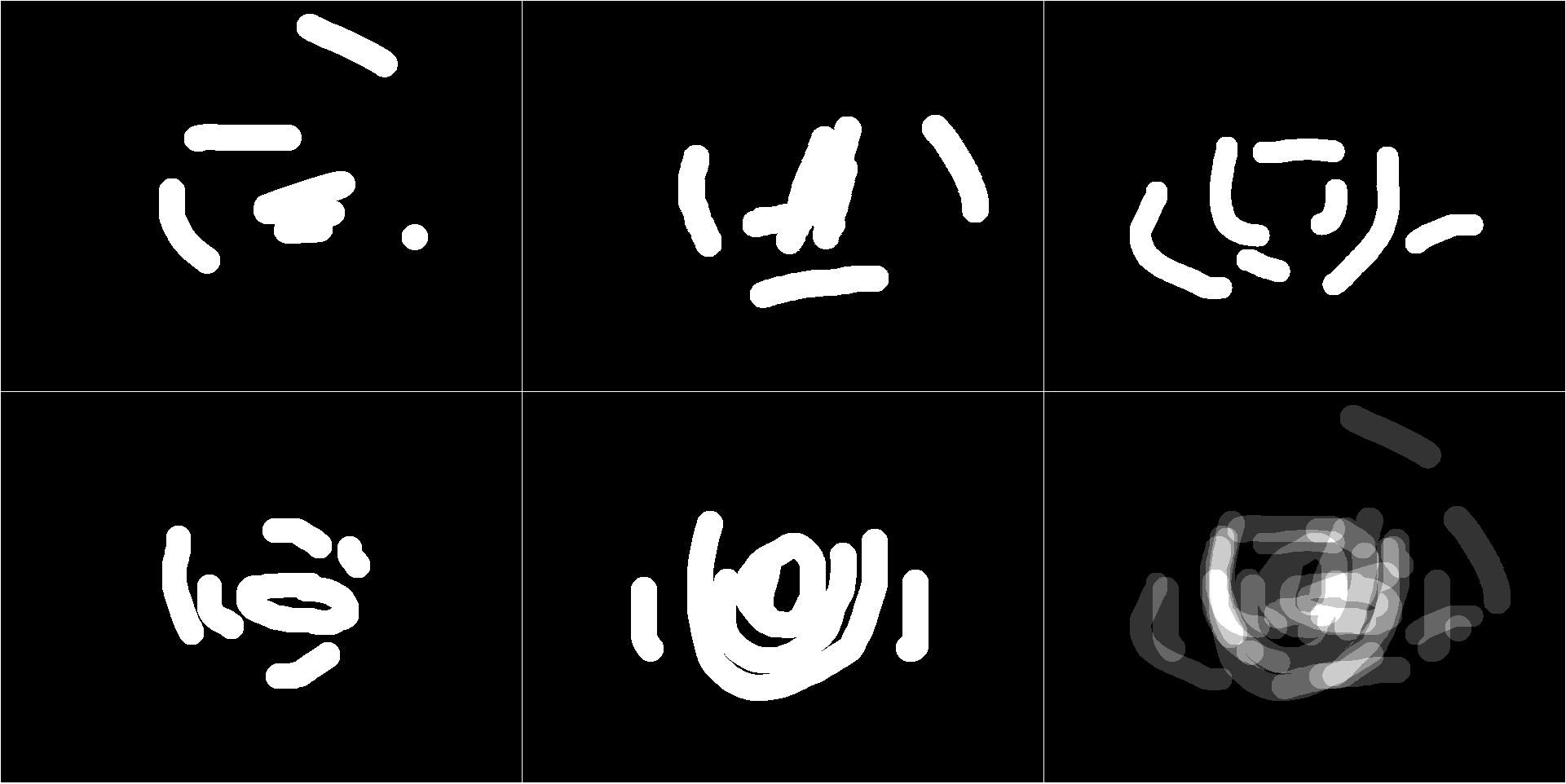}
            \caption{~}
        \end{subfigure}
        \hfill
        \begin{subfigure}[b]{0.218\textwidth}
          \centering
          \includegraphics[width=1\columnwidth]{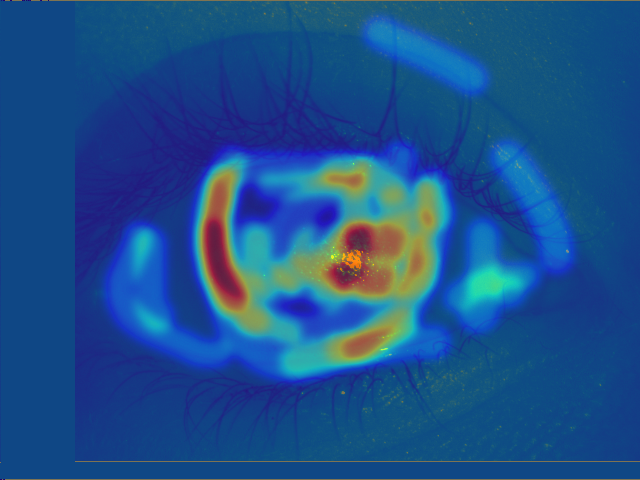}
          \caption{~}
        \end{subfigure}
        \hfill
        \begin{subfigure}[b]{0.218\textwidth}
          \centering
          \includegraphics[width=1\columnwidth]{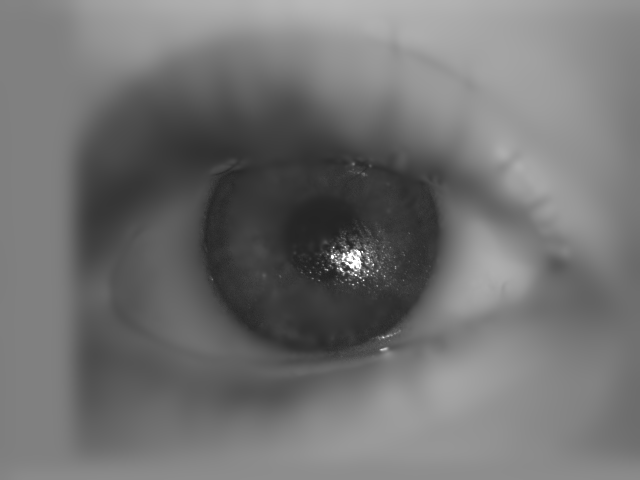}
          \caption{~}
        \end{subfigure}
    \end{subfigure}
    \caption{Creating human-guided training samples: the original image presented to human annotators (a); five individual correct annotations and a combination of those into the \textbf{saliency map} (b); a heatmap representation of the annotation density (c); the resulting training sample blurred locally with a magnitude (as defined in Sec. 4) inversely proportional to the annotation density (d).}
    \label{fig:incorp_process}
\end{figure*}

\noindent{\bf Superset of Available Iris PAD Datasets.}
We made an effort to acquire all known (to us) public research iris PAD datasets, to have a full representation of known presentation attack instruments (PAI),
ending up with more than 800,000 samples available. This collection was  first cleaned by removing duplicated 
and not ISO-compliant \cite{ISO_19794_6_2011} images. Second, it was split into {\bf training and validation} comprising 458,790 samples in 8 categories (live + seven PAIs), and a disjoint {\bf test set}, which is identical to the most recent LivDet-Iris competition benchmark \cite{Das_IJCB_2020} comprising 12,432 samples in six categories (live + 5 PAIs), as shown in Tab. \ref{tab:dataset}. This test dataset was excluded from all training and validation processes, and was withheld solely for final testing. This allows for comparison with the results of the LivDet-Iris 2020 competition, as well as providing an independently collected dataset to assess the generalization capabilities of the proposed approach.
Comprehensive descriptions of individual datasets can be found in the associated papers listed in Table \ref{tab:dataset}. Examples of PAIs available for this research, along with \textit{bona fide} samples are illustrated in Figure \ref{fig:attacks}, and individual dataset sample contribution statistics can be seen in the supplemental materials.
In this work the term \textit{abnormal} is assigned to the samples that differ from \textit{bona fide (live)} samples including presentation attacks.

\vskip1mm\noindent{\bf Human-Guided Region Saliency.} To facilitate human data collection, an online annotation tool was developed. Subjects were presented 8 types of images: bona fide and 7 abnormal types, as presented in Figure \ref{fig:attacks}.
Participants were not specifically trained in iris PAD or iris recognition tasks, and were associated with 
the University of Notre Dame
at the time of data collection.

On presentation of an iris image, users were asked to first select the type of image they believed it to be (one of eight types as above or \textit{unsure}). Next, 
users were asked to highlight at least five regions of the image supporting their decision. 
The regions highlighted were not constrained on size or on location within the image. The objective was to collect data on what information present in an ISO-compliant iris image leads humans (non-experts) to a classification decision. There are two reasons for using non-experts: (1) there are no experts formally trained in iris image examination (such experts do exist in, \eg fingerprint analysis); (2) to investigate whether a generalization boost can be obtained with help of non-experts in a given domain. The online annotation tool, with green highlights corresponding to user annotated regions, is presented in supplementary materials. 

Data from 150 subjects 
was collected in this experiment 
who annotated 30 unique image sets in total. Users were presented with an average of 27 images from one of the image sets. 
Image sets were assigned randomly to users 
and on average five subjects annotated each image. This simulates a scenario where some images have more annotation data available than others and hence the proposed approach must account for this imbalance.
Only annotations from correctly classified samples were kept. As PAD is a binary classification problem, it was deemed a correct classification if the subject selected correctly either a bona fide samples, or marked any of the 7 abnormal types as abnormal.

Note that merely collecting more labeled samples (bona-fide / abnormal) may (a) be impossible in the context of biometric attacks as these may be sparsely represented in datasets of ample size, and (b) not guide the network ``where to look'', opposite to the idea proposed in this paper. That is, the network, by simply observing more data, would still need to figure out relevant and irrelevant features without other guidance than the one through the loss function.

\begin{figure*}[t]
    \centering
    \includegraphics[width=1\linewidth]{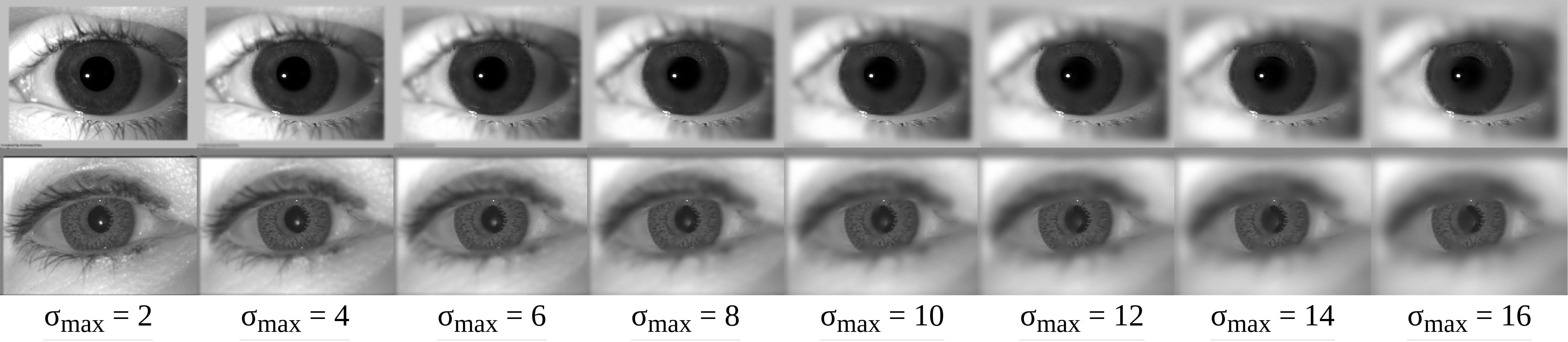}
    \caption{Illustration of annotations translating into saliency-encoded images for training. $\sigma_\mathrm{max}$  denotes the maximum blur is applied to unannotated regions. The top row is a bona fide iris and the bottom is a sample wearing a textured contact lens.
    } 
    \label{fig:blur}
\end{figure*}

The annotations are next used to create image representations called {\bf human saliency maps}, as shown in the bottom right image of Fig. \ref{fig:incorp_process}(b). 
Each correct annotation, as shown in five individual plots in Figure \ref{fig:incorp_process}(b), is weighted equally and combined.
The closer the pixel is to white in the saliency map, the more subjects selected that area as supporting their decisions. Black regions correspond to areas in which no subject annotated interesting features. These saliency maps provide the basis to incorporate the data into the training process, as outlined in the next Section \ref{sec:Incorporation}. 

\begin{table}[!ht]
 \centering
\caption{Human performance on the limited data used to construct saliency-encoded training images, measured in two scenarios: bona fide vs abnormal (independently of the abnormality type) and multi-class classification (indication of exact abnormality class was important).}
\footnotesize
\label{tab:statistics}
\begin{tabular}{|l|c|c|}
\hline
\textbf{Image Type} & \multicolumn{2}{c|}{{\bf Accuracy (\%)}} \\
& Bona fide / abnormal & Exact type\\ \hline\hline
Bona fide                                                                 & 56.3                   & 56.3 \\ \hline
Textured contact lenses & 65.13                  & 32.89 \\\hline
Paper printouts                                                          & 94.27                  & 64.53 \\ \hline
Post-mortem                                                               & 98.0                   & 79.43 \\ \hline
Synthetic                                                                 & 83.81                  & 49.0 \\ \hline
Artificial                                                                & 70.19                  & 34.82 \\ \hline
Textured contact   lenses printed                                                     & 88.17                  & 27.46 \\\hline
Diseased                                                     & 77.32                  & 47.73 \\\hline
\end{tabular}
\vskip-3mm
\end{table}

Classification accuracy from the human annotators is shown in Table \ref{tab:statistics}.
The most difficult type to classify correctly is textured contact lenses. This might be due to the fact that lens manufacturers design them to  mimic genuine patterns. Conversely, the highest human classification accuracy (98\%) was for post-mortem samples. The presence of metal retractors to hold the eyelids open
potentially contributes to this. Human annotators also performed well on paper printouts, possible due to well-visible pattern spread through the images. 
Interestingly,
humans not trained in iris recognition, and classifying near-infrared iris samples, tend to have lower accuracy for bona fide samples.

\section{Saliency-Encoded Training Images}
\label{sec:Incorporation}

Given a set of human-annotated iris samples, 
various levels of 
Gaussian filtering are applied
to de-emphasize regions not marked as salient by humans.
The intuition 
is that human annotators 
are able to restrict attention to regions relevant to the decision,
as opposed to features of the training samples that may have incidental correlation with class labels. 
The magnitude of the blur 
(Gaussian kernel width $\sigma$)
has a simple relation 
to how frequently a given region was annotated, and in particular
regions selected by the largest number of annotators remain unchanged. 
We use blurring rather than binary masks due to (a) the need of gradual information suppression reflecting the human saliency,
and (b) sharp edges around a binary mask could constitute ``fake'' image features that impact training. This approach mitigates these sharp edges between edges by applying Gaussian blur of $\sigma = 5$ to the saliency map. 

An important degree of freedom 
is the maximum strength of blurring $\sigma_\mathrm{max}$ applied to non-annotated regions, serving as baseline when calculating all remaining annotation map-dependent blur levels $\sigma$ for a given sample. Instead of making arbitrary choices about $\sigma_\mathrm{max}$, we use a combination of $\sigma_\mathrm{max} \in \{2,4,6,8,10,12,14,16\}$ as particular levels of blur ``aggressivness'' may make more sense for some abnormal types, or some values are better than others across all abnormal types. Thus, all annotated regions are blurred with a blur level $\sigma$ based on a function of $\sigma_\mathrm{max}$ and the number of subjects that highlighted a specific region: 
$$
\sigma = (\sigma_\mathrm{max}(1-\rho))^4 / \sigma_\mathrm{max}^3
$$

\noindent
where $\rho$ is the fraction between 0 and 1 of annotators that selected a given image area. Thus, if zero subjects highlight a region ($\rho=0$) then $\sigma=\sigma_\mathrm{max}$. And if all annotators working on a given image marked that region as important to them, then $\sigma=0$ and these regions will be passed to the network unchanged. Figure \ref{fig:blur} illustrates how the human annotation maps translate to the human-aided training data for various 
$\sigma_\mathrm{max}$. 
Increasing $\sigma_\mathrm{max}$ subjects unannotated regions to stronger blurring.

This straightforward 
mechanism of guiding the model to learn more human-like decisions has never been explored before. The next section will demonstrate its effectiveness in the case of limited training data, which is particularly relevant for biometric presentation attack detection.

\section{Experiments}

\noindent{\bf Setup.} We selected D-NetPAD \cite{Sharma_IJCB_2020} as the most recent, open-source deep-learning-based iris PAD algorithm, demonstrating good results on the LivDet-Iris 2020 benchmark \cite{Das_IJCB_2020}. 
To ensure that the results presented in the evaluation section are the result of the application of human data, and not due to parameter optimization or modification to the method, no changes were made to the model parameters from the publicly available code.
That is, the learning rate was set at $0.005$, batch size was $20$ and the number of epochs was $50$ for all experiments in this section. SGD with a momentum of $0.9$ was used as the optimization algorithm and cross-entropy loss function was applied. No additional train time image augmentations were applied. 
All data is segmented (a SegNet-based method \cite{Trokielewicz_IVC_2020}), cropped, resized to $224\times224$ and used as input.

\vskip1mm\noindent{\bf Leave-one-type-out experiments.} Increased generalization means that the model can better classify new types of inputs unseen in training. This perfectly fits into biometric presentation attack detection, where -- realistically -- we cannot assume that only abnormal types present in the training data will be observed during testing.  
To evaluate effectiveness of the proposed method, seven ``leave-one-abnormal-type-out'' experiments are run.
For each experiment, one abnormal type 
is omitted from both training and validation, and present only in testing. As this is a binary classification problem, bona fide samples (yet from disjoint sets) are used in all training, validation and test sets. In each of the 7 experiments, the left-out class represents an unknown type of abnormality for the model, and  
the performance on these left-out sets is indicative of the model's generalization capability. Because the \textit{textured contacts \& printout} type contains information about both contacts and printouts, these samples were excluded from training and validation for experiments with the \textit{textured contacts} and \textit{paper printouts}, thus maintaining complete model ignorance of the nature of the test data. Similarly, \textit{textured contacts} and \textit{paper printouts} were excluded from the experiment where \textit{textured contacts \& printouts} was the held-out test set. 
The models trained in the leave-one-abnormal-type-out scenario are then tested on unseen bona fide samples taken from the LivDet-Iris 2020 dataset, and unseen abnormal samples originating from the large dataset. Taking bona fide samples from LivDet-Iris 2020 corpus maximizes the ``open-setness'' of these  experiments, as these bona fide samples were acquired independently of all the training and validation data shown in Table \ref{tab:dataset}.

\begin{figure*}[t]
  \begin{subfigure}[b]{1\textwidth}
      \begin{subfigure}[b]{0.24\textwidth}
          \centering
          \includegraphics[width=1\columnwidth]{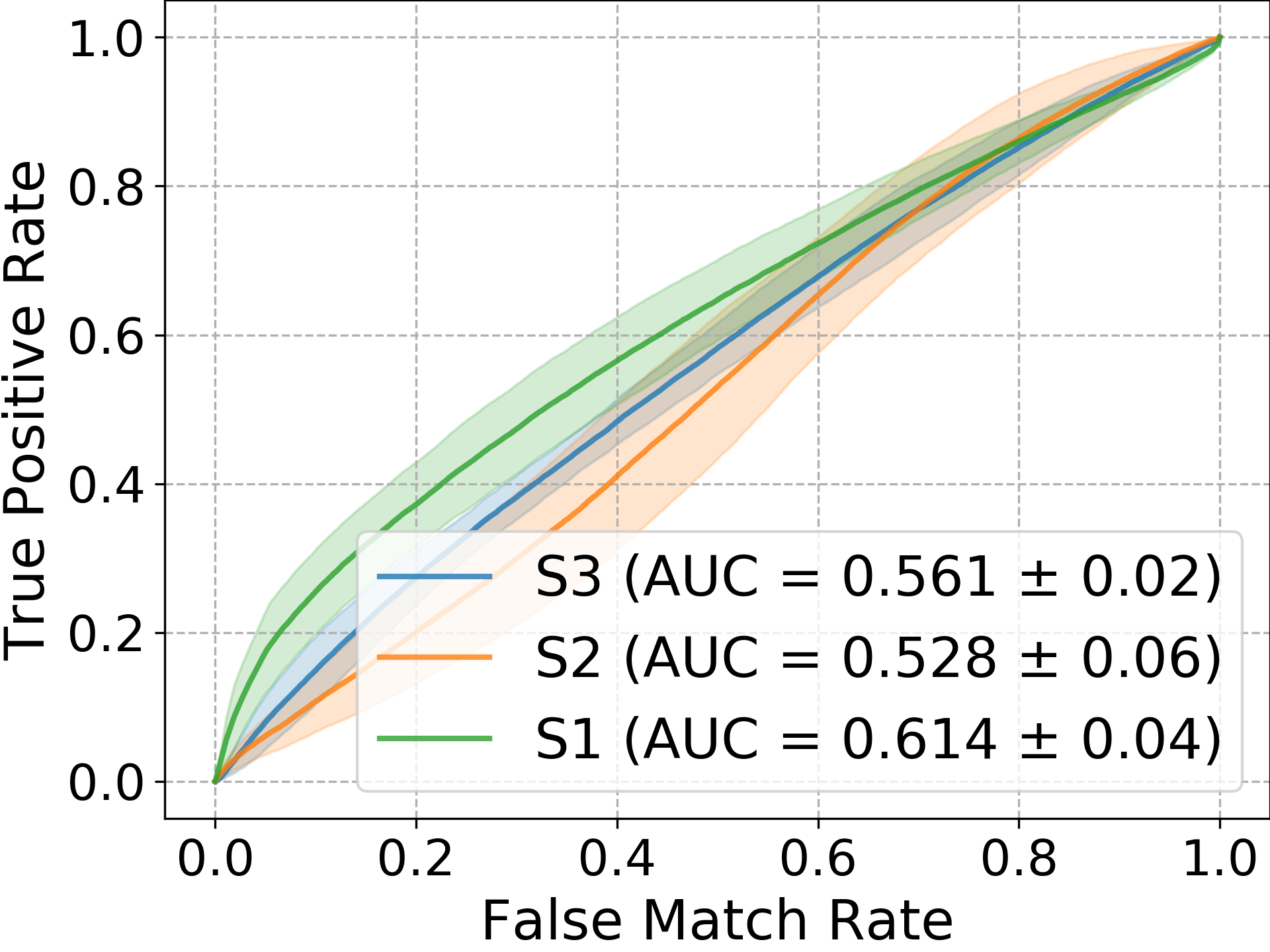}
          \vskip-2mm
          \caption{Textured contact lens}
      \end{subfigure}
      \hfill
      \begin{subfigure}[b]{0.24\textwidth}
          \centering
          \includegraphics[width=1\columnwidth]{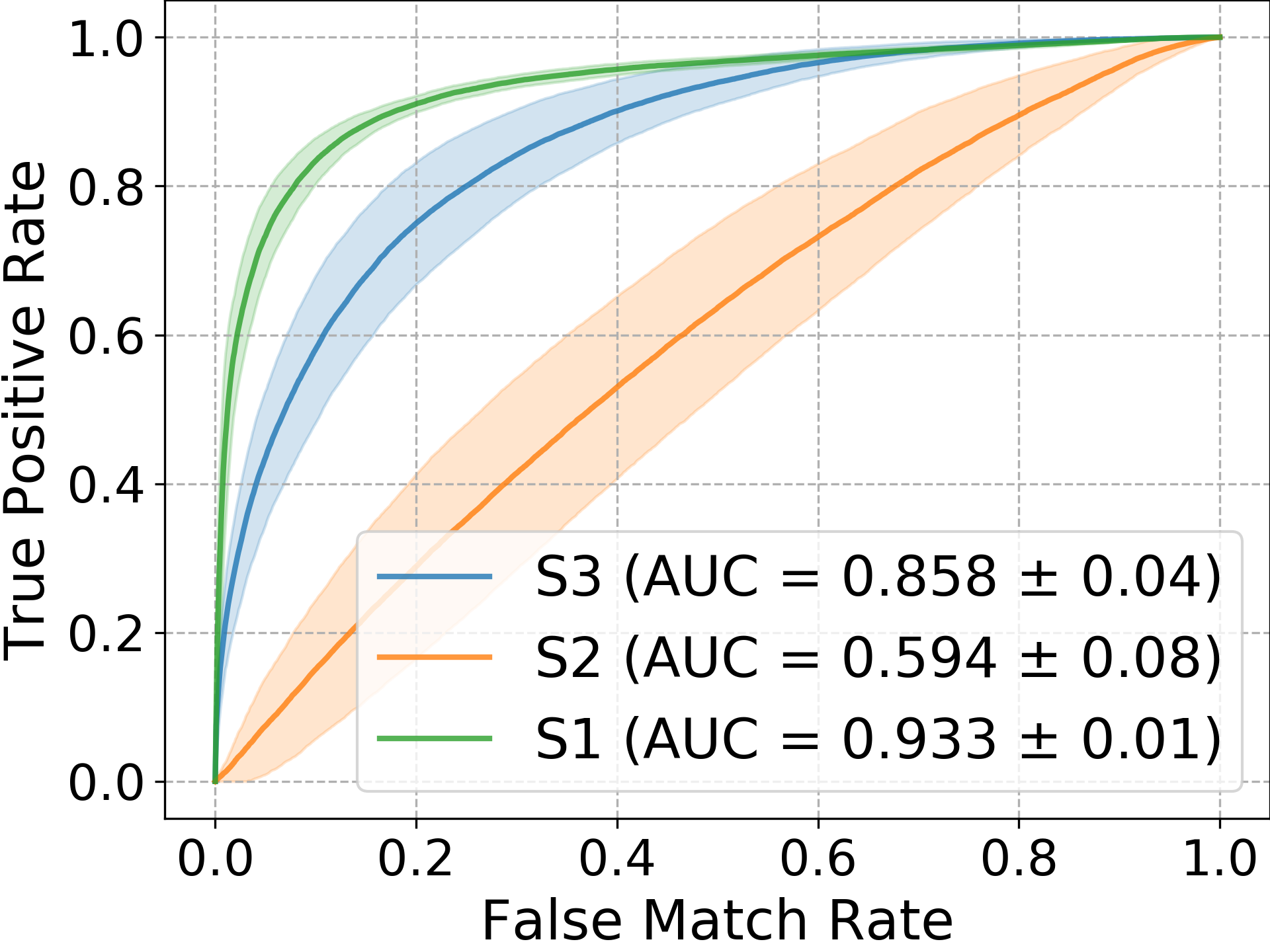}
          \vskip-2mm
          \caption{Paper printouts}
      \end{subfigure}
      \hfill
      \begin{subfigure}[b]{0.24\textwidth}
          \centering
          \includegraphics[width=1\columnwidth]{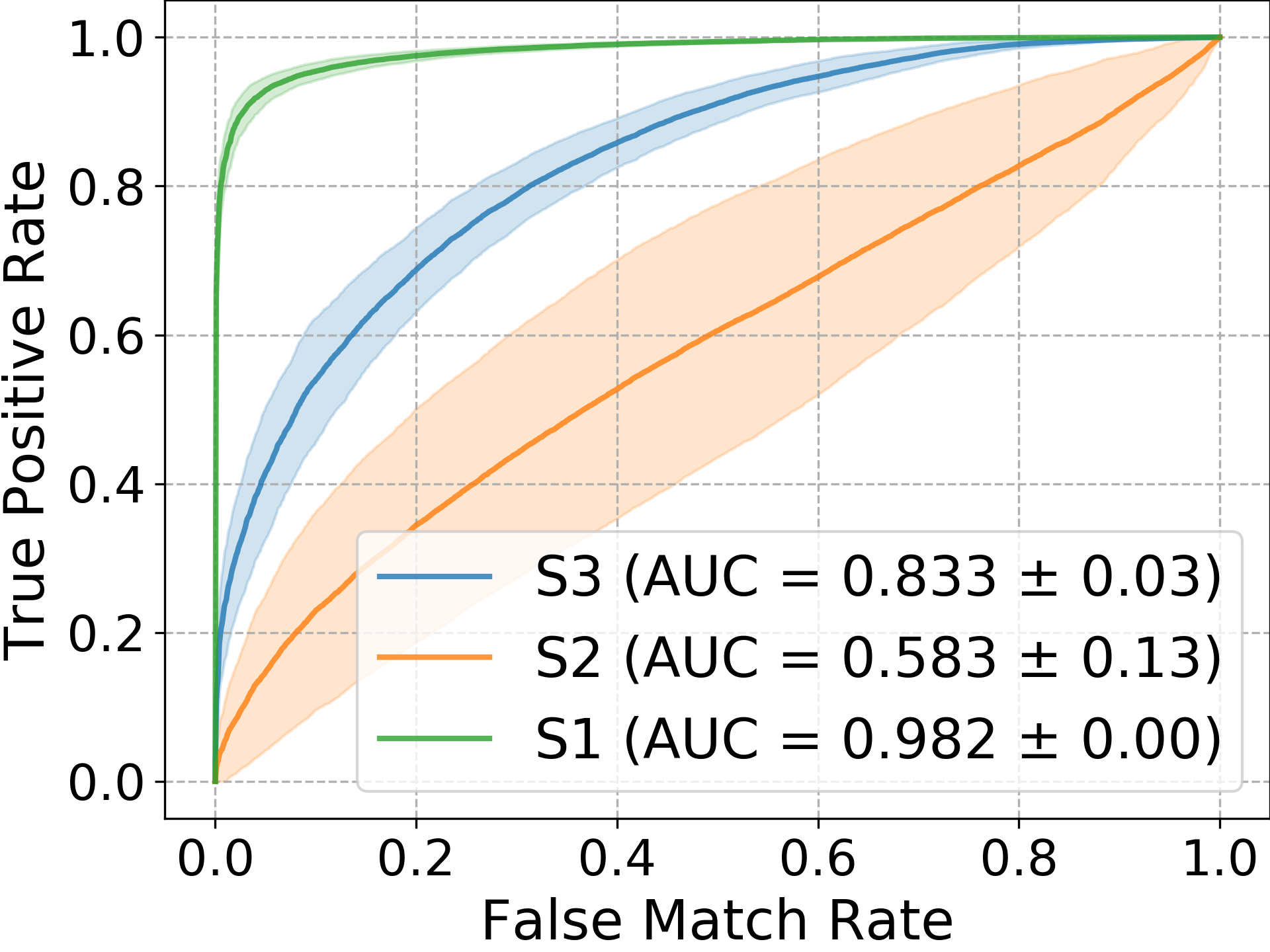}
          \vskip-2mm
          \caption{Post-mortem}
      \end{subfigure}
      \begin{subfigure}[b]{0.24\textwidth}
          \centering
          \includegraphics[width=1\columnwidth]{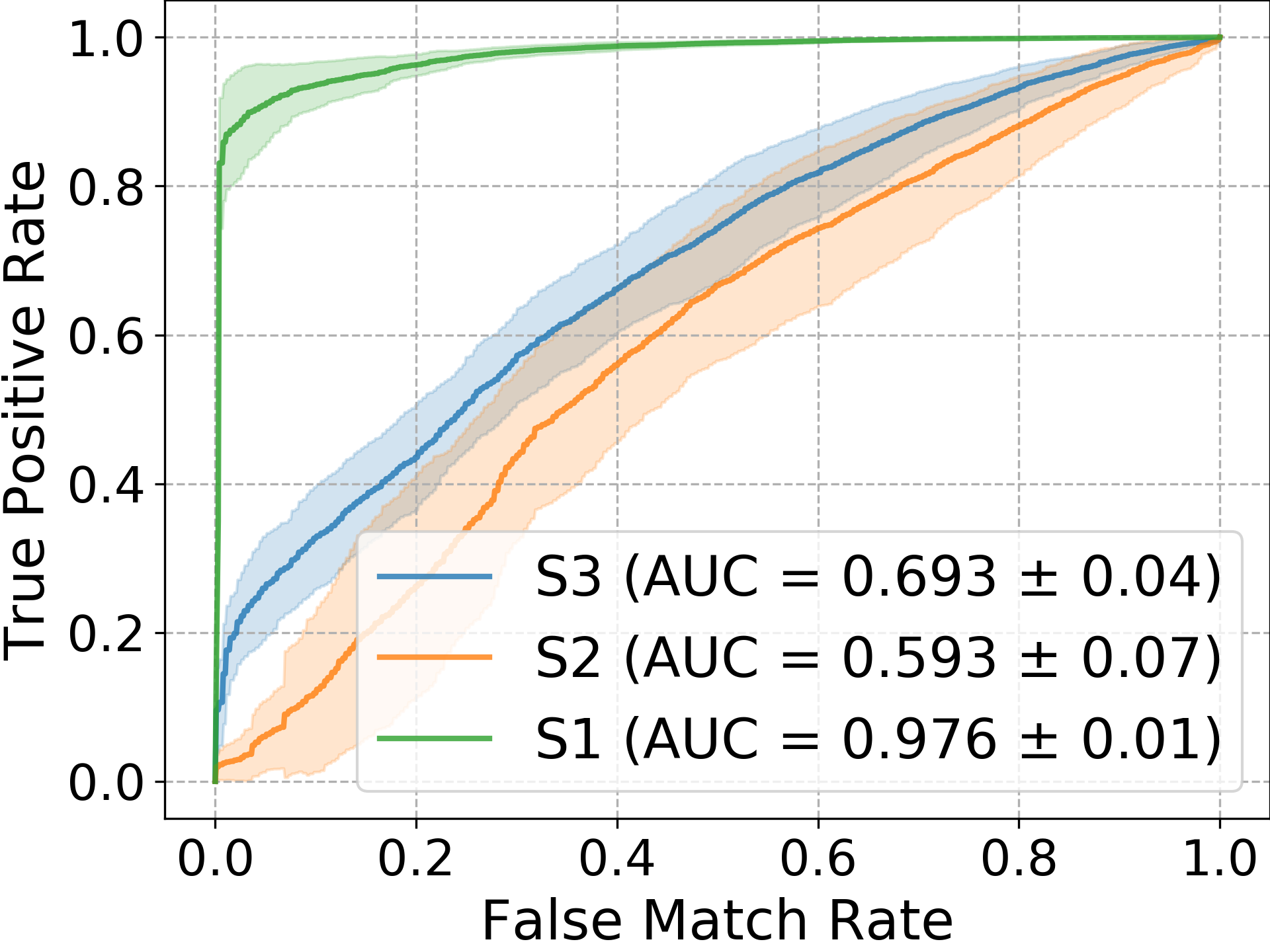}
          \vskip-2mm
          \caption{Artificial}
      \end{subfigure}
  \end{subfigure}\vskip3mm
  \begin{subfigure}[b]{1\textwidth}
      \centering
      \begin{subfigure}[b]{0.24\textwidth}
          \centering
          \includegraphics[width=1\columnwidth]{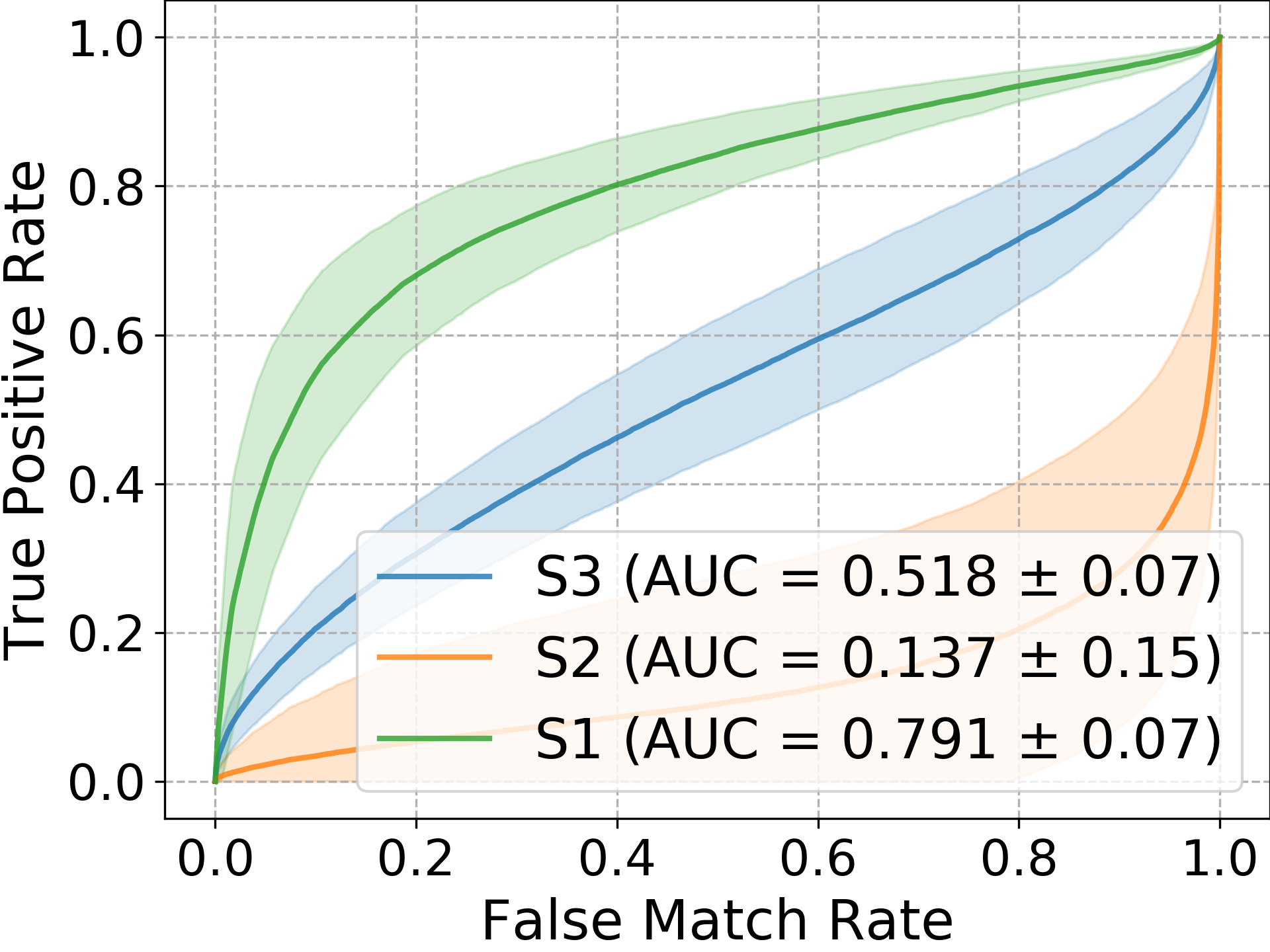}
          \vskip-2mm
          \caption{Synthetic}
      \end{subfigure}
      \begin{subfigure}[b]{0.24\textwidth}
          \centering
          \includegraphics[width=1\columnwidth]{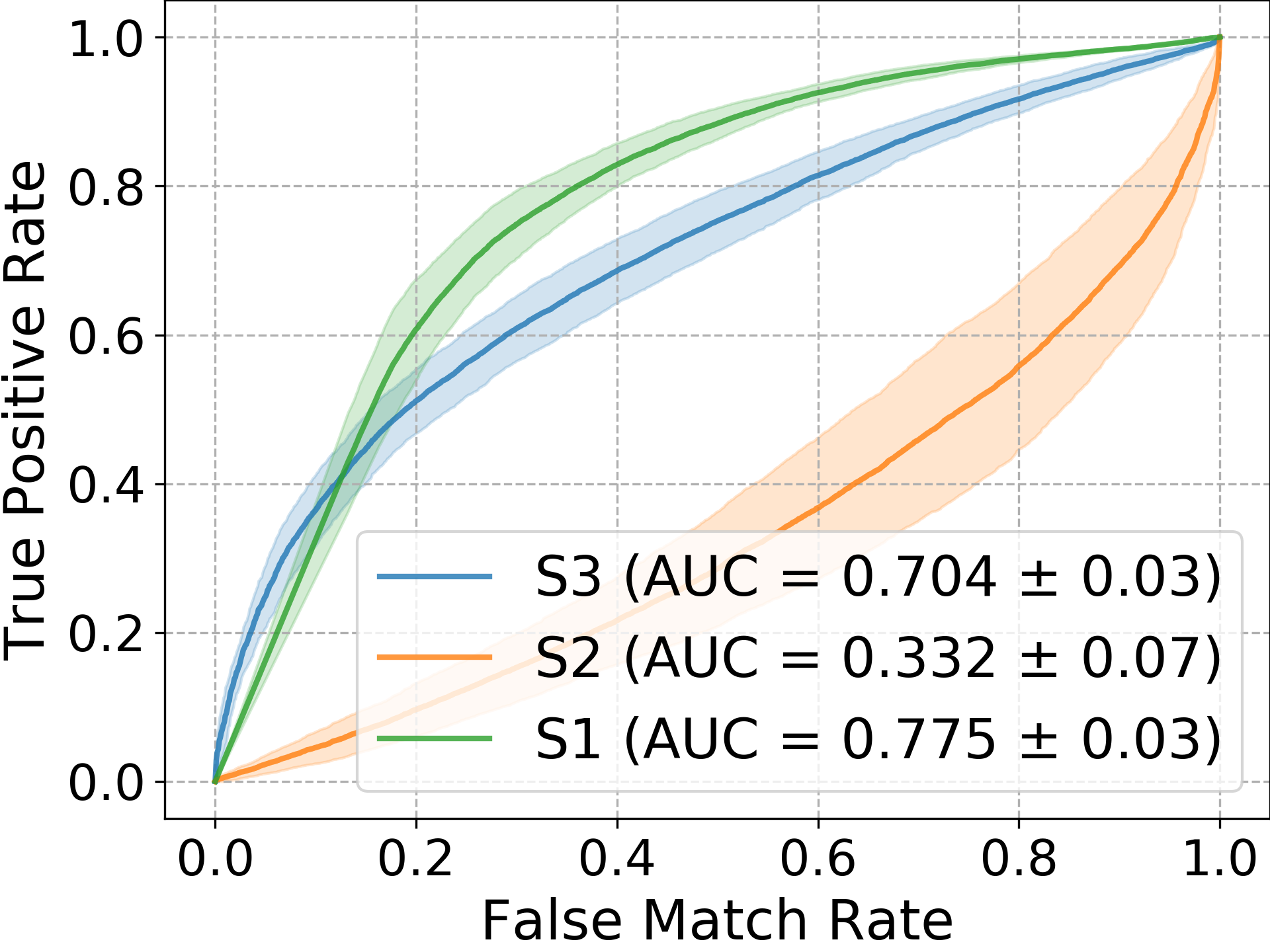}
          \vskip-2mm
          \caption{Diseased}
      \end{subfigure}
      \begin{subfigure}[b]{0.24\textwidth}
          \centering
          \includegraphics[width=1\columnwidth]{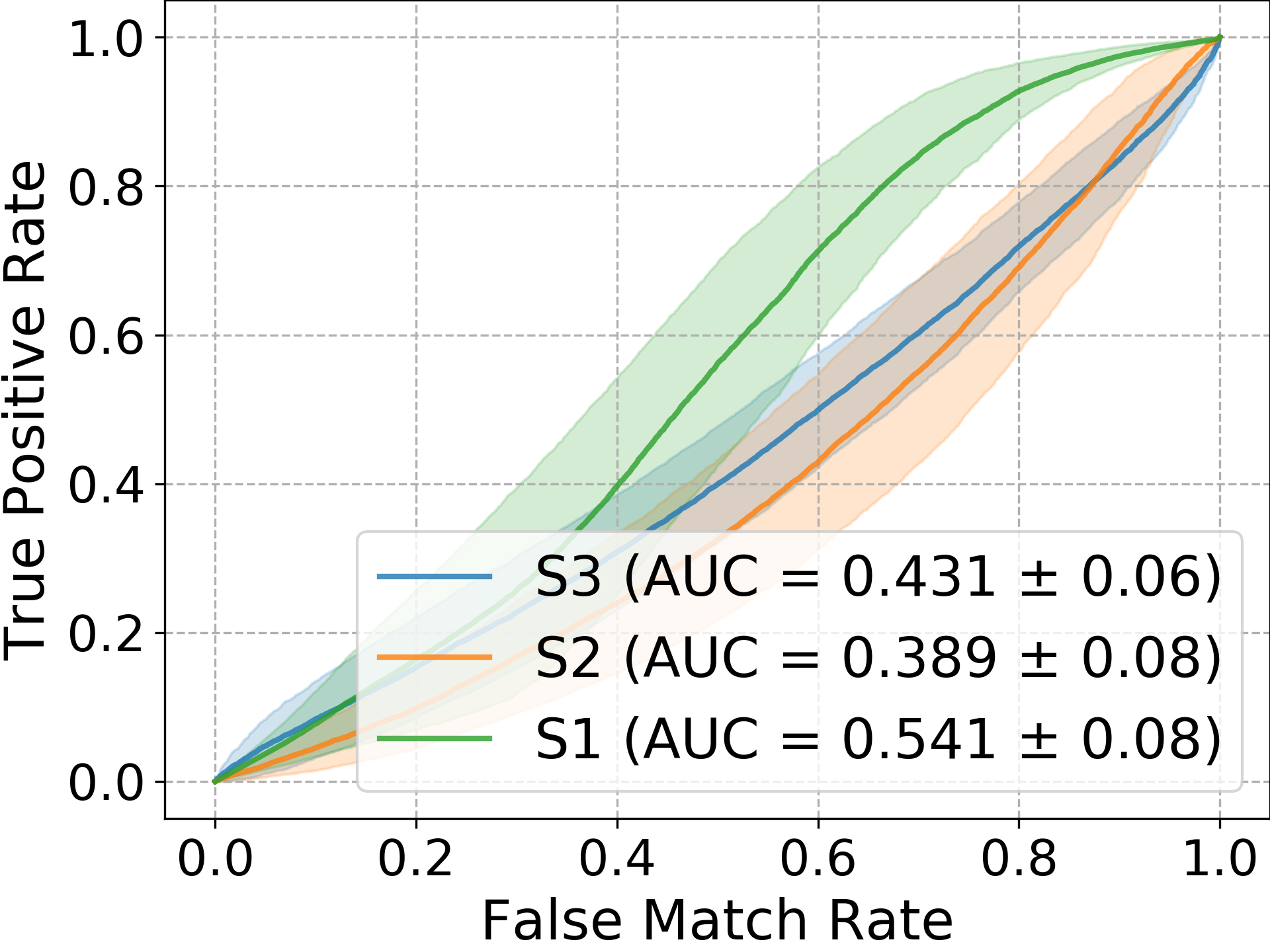}
          \vskip-2mm
          \caption{Contact lens \& printed}
      \end{subfigure}
      \label{fig:roc_bestlayers}
  \end{subfigure}
  \caption{ROC curves in leave-one-abnormal-type-out experiments obtained in three training scenarios. The left-out type is named in the plot caption. Shaded bands represent $\pm 1$ standard deviation along the TPR axis obtained in repeated experiments. \textcolor{s1}{\bf S1} refers to scenario 1 using the large dataset in training. \textcolor{s2}{\bf S2} refers to limited training dataset with \textbf{no} human salience info incorporated. \textcolor{s3}{\bf S3} refers to the training 
  with human-salience-encoded versions of the same original images as \textcolor{s2}{S2}. In all cases, there is a significant performance increase when human salience info is used in training (compare \textcolor{s3}{S3} with \textcolor{s2}{S2}).}
  \label{fig:roc}
\end{figure*}

\vskip1mm\noindent{\bf Evaluation scenario S1: Training on the Large Dataset.} 
In this scenario, models are trained on the unmodified training data described in Table \ref{tab:dataset}, starting from pre-trained ImageNet weights, as common practice suggests \cite{boyd2019deep}. Before model training, all samples that have corresponding correct human annotations ($765$ in total) are removed from the training and validation sets. This means the human-annotated data used in other scenarios was \textit{unseen}. 
There are many more bona fide samples available than abnormal samples, as shown in Tab. \ref{tab:dataset}. To prevent the over-representation of one class in training, all abnormal samples are taken along with a randomly sampled set of bona fide samples equal to the number of abnormal samples. 
Models trained in this scenario are referred to as the {\bf large data models}, achievable in the situation of having a ``large-enough'' training dataset, which usually is not the case in biometric presentation attack detection.

\vskip1mm\noindent{\bf Evaluation scenario S2: Training on Regular Limited Data.} In this scenario, the same ImageNet-initiated models are tuned with $765$ live and abnormal samples minus the samples corresponding to the left-out type, for which correct human annotations were collected (but {\bf not} used here). Additionally, these images are augmented by applying a Gaussian blur $\sigma \in \{2,4,6,8,10,12,14,16\}$ to the entire image and combined with the unblurred versions of the samples, increasing the size of the dataset nine-fold. This is to ensure that the results obtained for human-annotated training data (scenario S3 below) are due to the model learning human-aided features, not the addition of the global blur as image augmentations. 
This scenario simulates a situation where limited dataset representing a given domain is available, and human-aided augmentations are not used. 
Models trained according to this scenario will be referred to as the {\bf limited data models}.

\vskip1mm\noindent{\bf Evaluation scenario S3: Training on Human-Aided Limited Data.} The same training as scenario S2 is applied, but using human saliency encoded versions of all $765$ training samples, again minus samples corresponding to the left-out type. Networks are trained on a combined set comprised of images transformed with all maximum blur levels, as described in Sec. \ref{sec:Incorporation}. Samples with no blur are not included in this scenario as they were in S2. 
Models trained according to this scenario are referred to as the {\bf human-aided models}.

\vskip1mm\noindent{\bf Common settings across all scenarios.} The validation set used for best epoch selection was the same in all scenarios, and the validation images were not blurred. Since all models 
accept cropped and resized images as input, 
both original and the user-annotated images were cropped in the same way. 
As with the large training dataset, the common validation set is also balanced such that the number of bona fide and abnormal samples is equal.

\section{Evaluation Results}
\begin{table*}[!t]
 \centering
 \begin{threeparttable}
 \footnotesize
\centering
\caption{LivDet-Iris 2020 competition results (in \%) compared to the proposed approaches.}
\label{tab:livdet_results}
\begin{tabular}{|*{10}{c|}}
\hline
\textbf{Method} & \multirow{2}{*}{\textbf{Algorithm}} & \multicolumn{5}{c|}{\textbf{APCER}}  & \multicolumn{2}{c|}{ \textbf{Overall Performance}} & \multirow{2}{*} {\textbf{ACER}} \\
\cline{3-9}
\textbf{category} &  & \textbf{PP} & \textbf{CL} & \textbf{DP} & \textbf{AR} & \textbf{PM} & \textbf{APCER$_{\mbox{\footnotesize average}}$} & \textbf{BPCER} & \\
 \hline\hline
 \multirow{3}{*}{\makecell{\textbf{Livet Iris 2020}\\{ \textbf{Submissions}}}} &  Team: USACH/TOC & 23.64 & 66.01 & 9.87  & 25.69 & 86.10 & 59.10 & 0.46 & 29.78 \\
 \cline{2-10}
  &Team: FraunhoferIGD & 14.87 & 72.80 & 53.08 &  19.04 & 0 & 48.68 & 11.59 & 30.14 \\
 
  \cline{2-10}
  & Competitor-3 & 72.64 & 43.68 & 83.95 & 73.19 & 89.85 & 57.8  & 40.31 & 49.06 \\
    \hline
    \hline
\multirow{3}{*}{\textbf{This Work}}  &  Large Training Data (S1) & 9.34 & 32.89 & 3.70 & 2.03 & 0.55 & 21.74 & 0.47
  & 11.1 \\ 
\cline{2-10}
 & Limited Training Data (S2) & 1.91 & 5.05 & 1.23 & 3.512 & 0.09 & 3.66 & 93.08 & 48.37\\
\cline{2-10}
& {\bf Human-Aided (S3)} & 9.06 & 36.65 & 0.0 & 2.77 & 1.37
& 24.14 & 8.61 & {\bf 16.37} \\
\hline
\end{tabular}
\begin{tablenotes}
      \footnotesize
      \item \textbf{PP:} Paper printouts; \textbf{CL:} Textured contacts ; \textbf{DP:} Display; \textbf{AR:} Artificial; \textbf{PM:} Post-mortem; \textbf{APCER:} Attack Presentation Classification Error Rate (abnormal called bona fide); \textbf{APCER$_{\mbox{\footnotesize average}}$:} APCER averaged across all attack types; \textbf{BPCER:} Bonafide Presentation Classification Error Rate (bona fide called abnormal); {\bf ACER:} average of BPCER and APCER$_{\mbox{\footnotesize average}}$.
    \end{tablenotes}
    \end{threeparttable}
\end{table*}

\begin{figure*}[!ht]
  \begin{subfigure}[b]{1\textwidth}
      \begin{subfigure}[b]{0.24\textwidth}
          \centering
          \includegraphics[width=1\columnwidth]{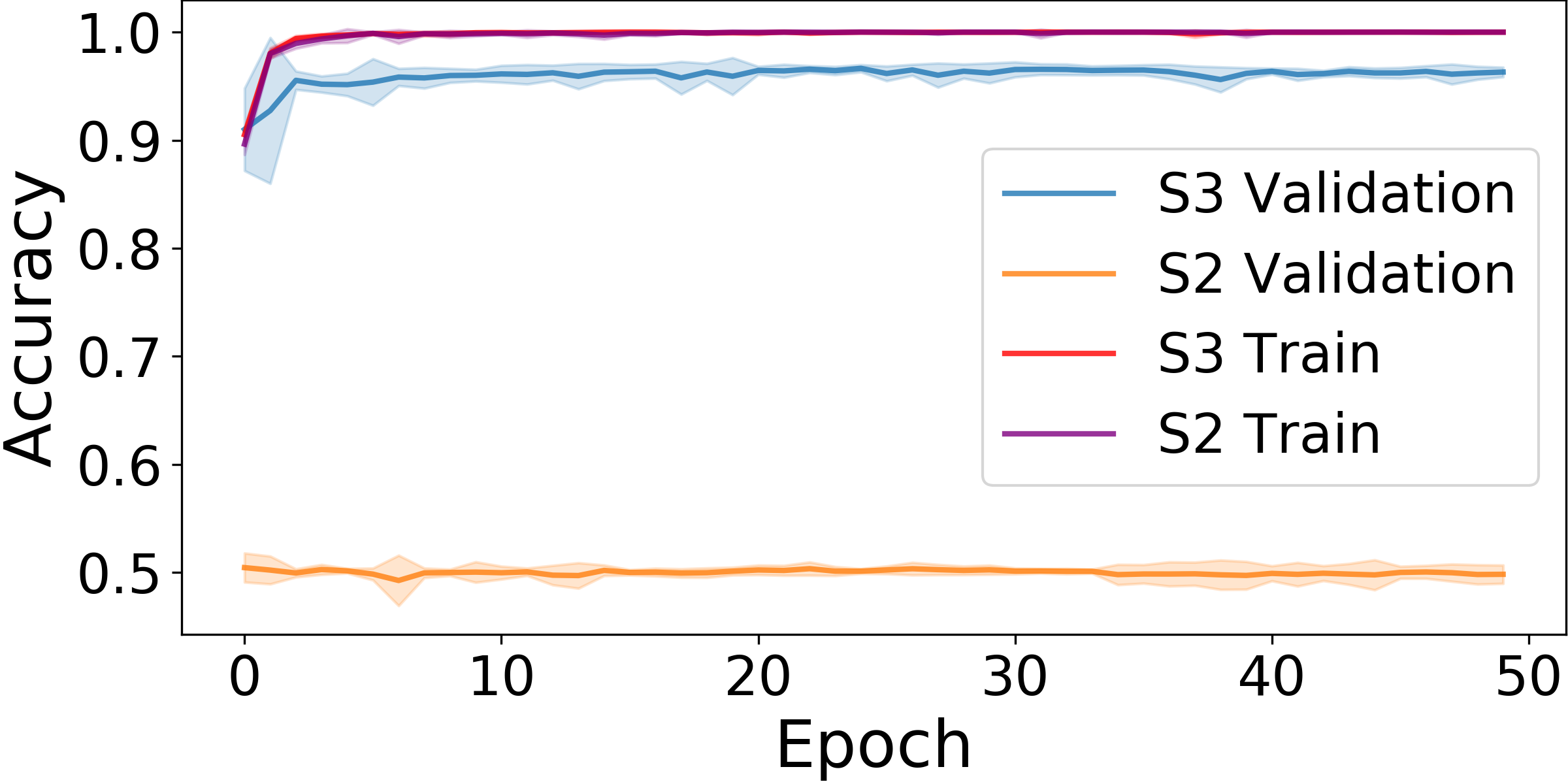}
          \vskip-2mm
          \caption{Textured contact lens}
      \end{subfigure}
      \hfill
      \begin{subfigure}[b]{0.24\textwidth}
          \centering
          \includegraphics[width=1\columnwidth]{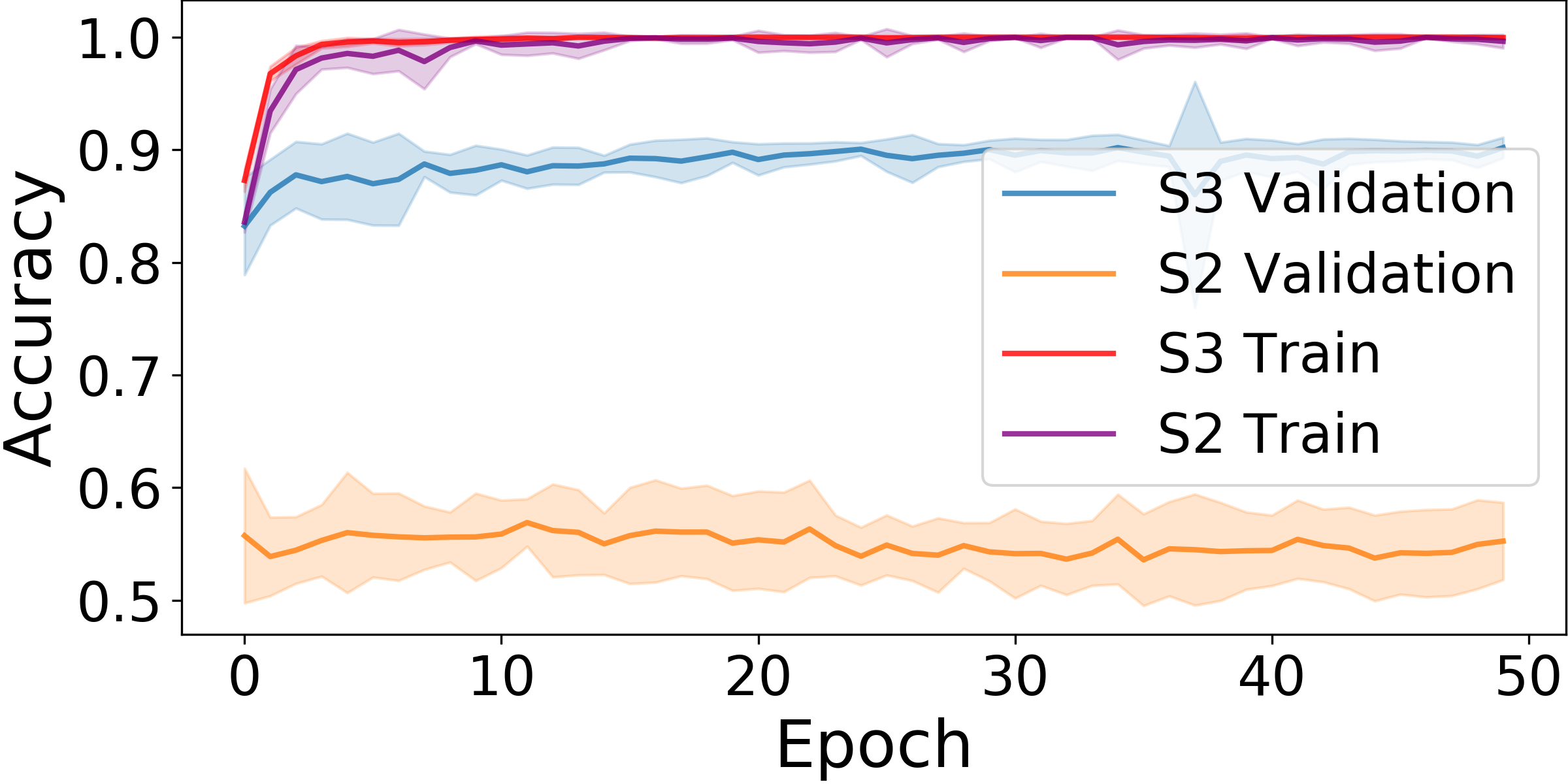}
          \vskip-2mm
          \caption{Paper printouts}
      \end{subfigure}
      \hfill
      \begin{subfigure}[b]{0.24\textwidth}
          \centering
          \includegraphics[width=1\columnwidth]{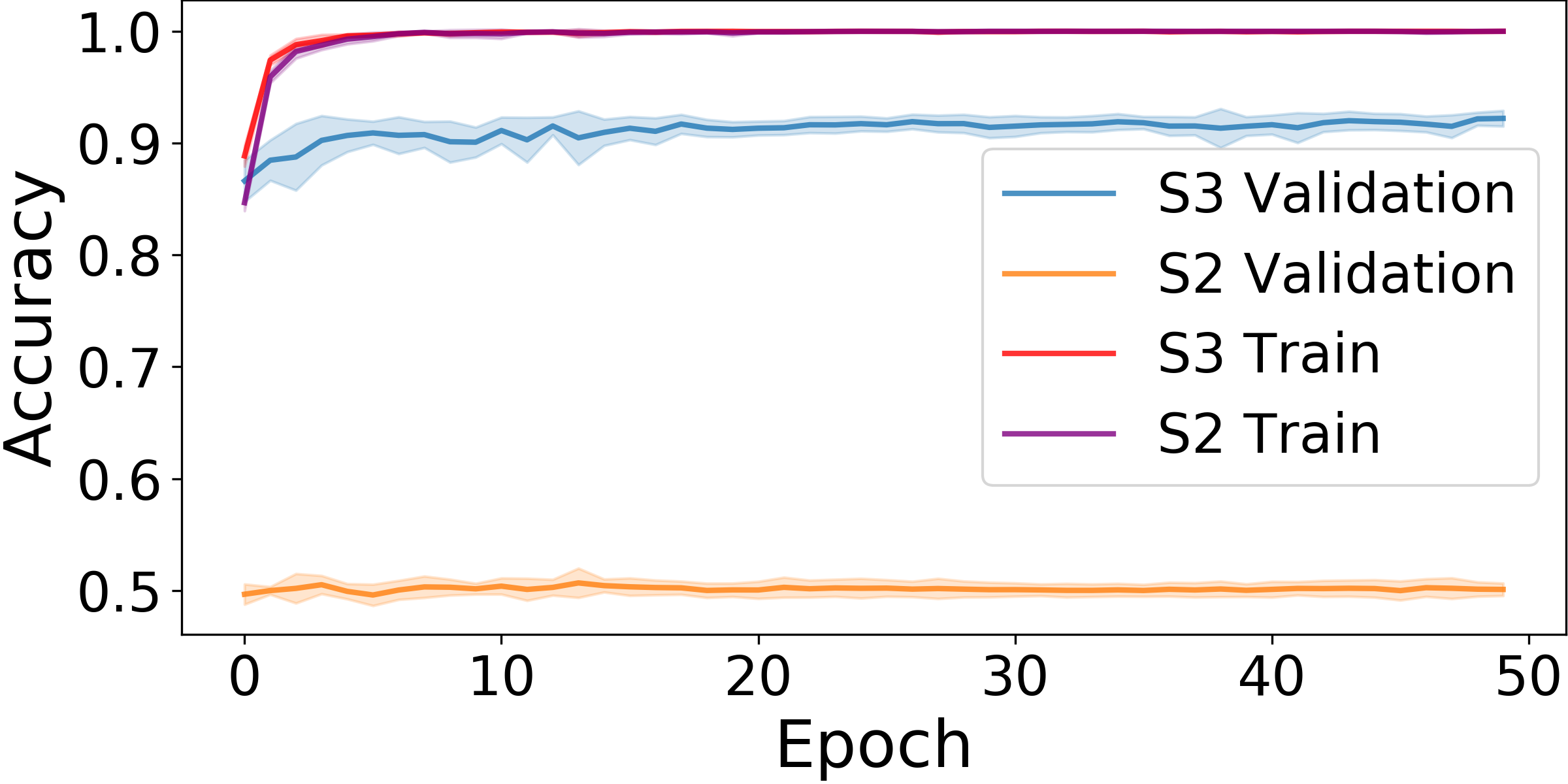}
          \vskip-2mm
          \caption{Post-mortem}
      \end{subfigure}
      \begin{subfigure}[b]{0.24\textwidth}
          \centering
          \includegraphics[width=1\columnwidth]{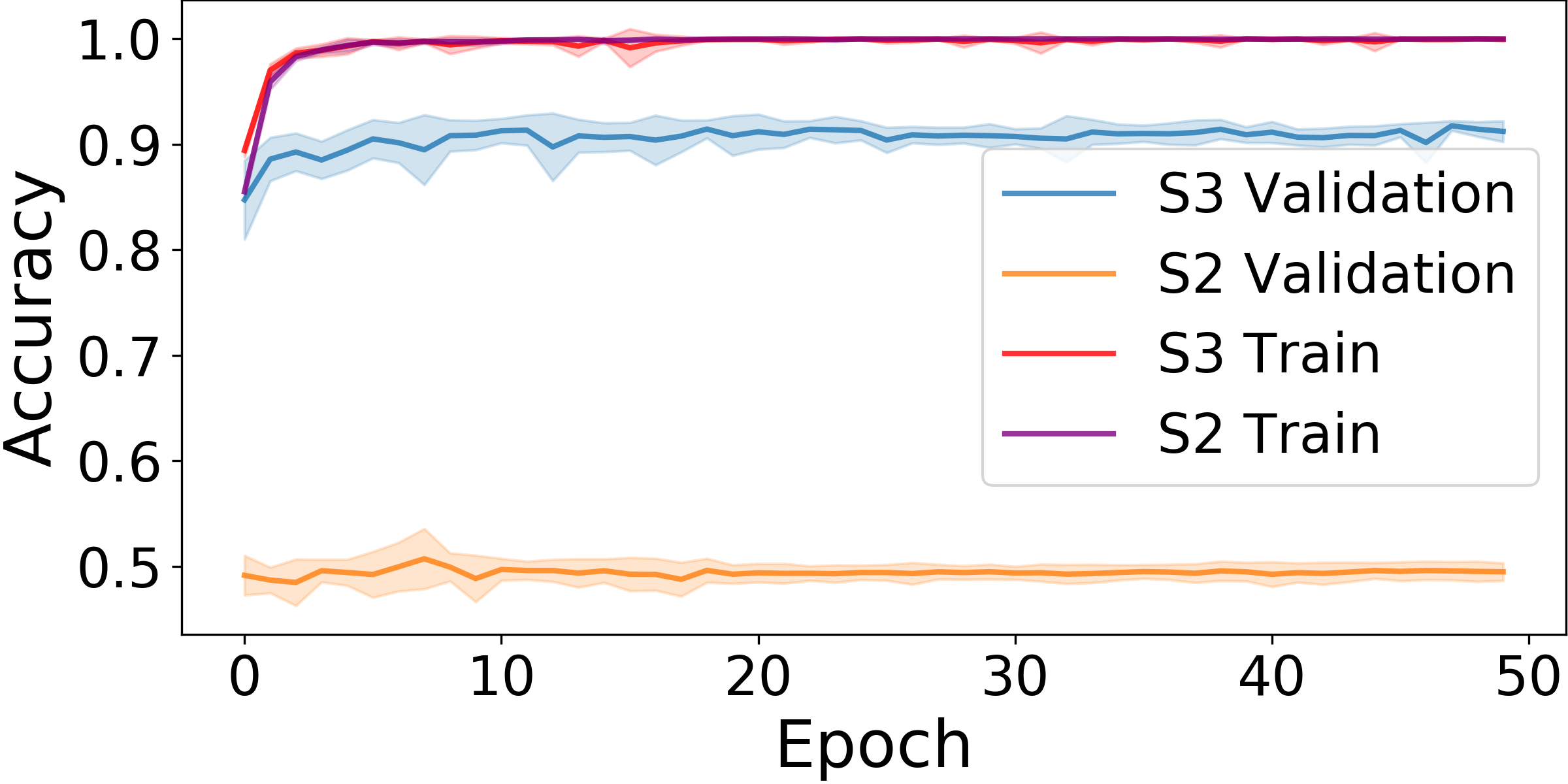}
          \vskip-2mm
          \caption{Artificial}
      \end{subfigure}
  \end{subfigure}\vskip3mm
  \begin{subfigure}[b]{1\textwidth}
      \centering
      \begin{subfigure}[b]{0.24\textwidth}
          \centering
          \includegraphics[width=1\columnwidth]{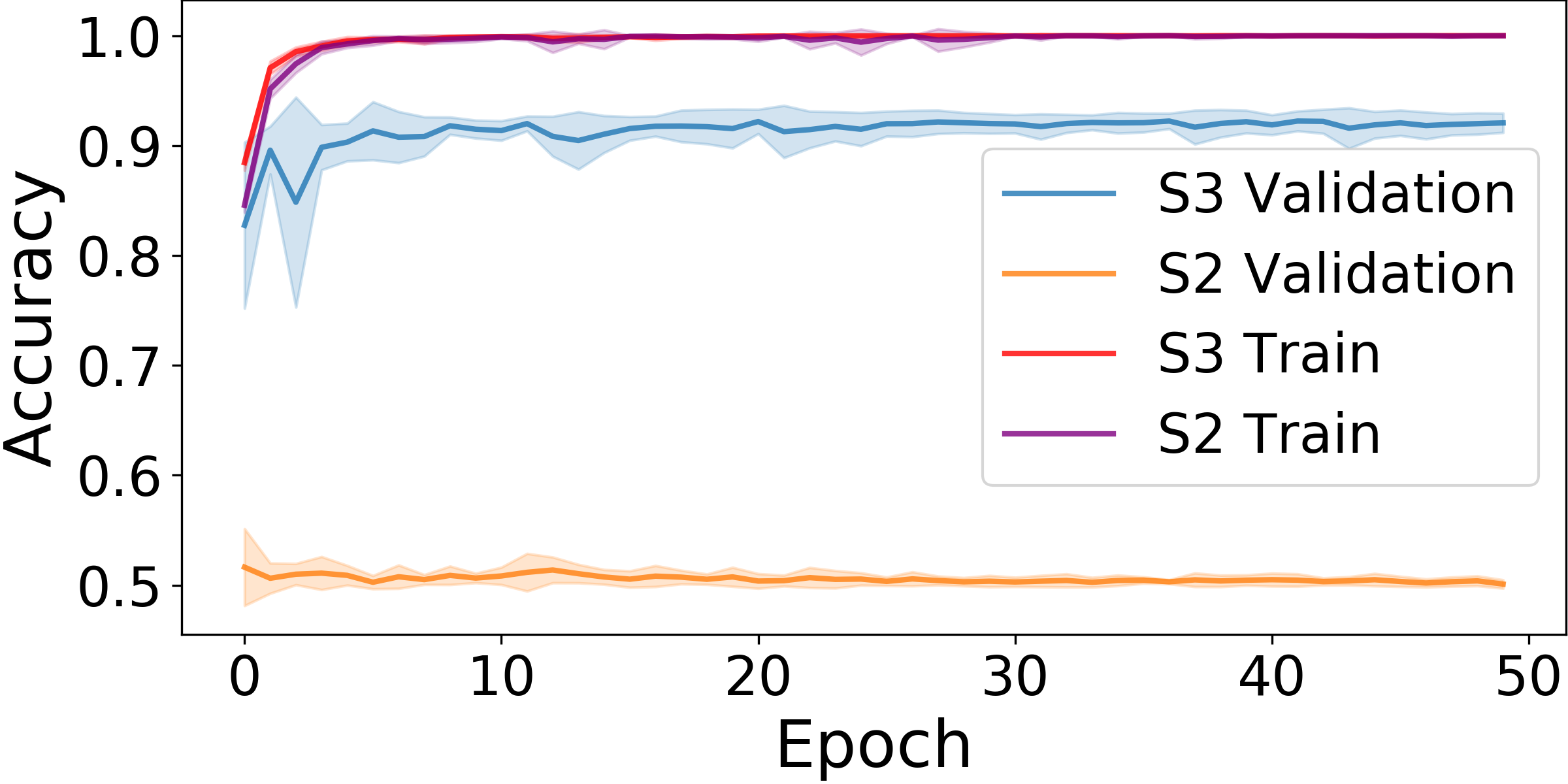}
          \vskip-2mm
          \caption{Synthetic}
      \end{subfigure}
      \begin{subfigure}[b]{0.24\textwidth}
          \centering
          \includegraphics[width=1\columnwidth]{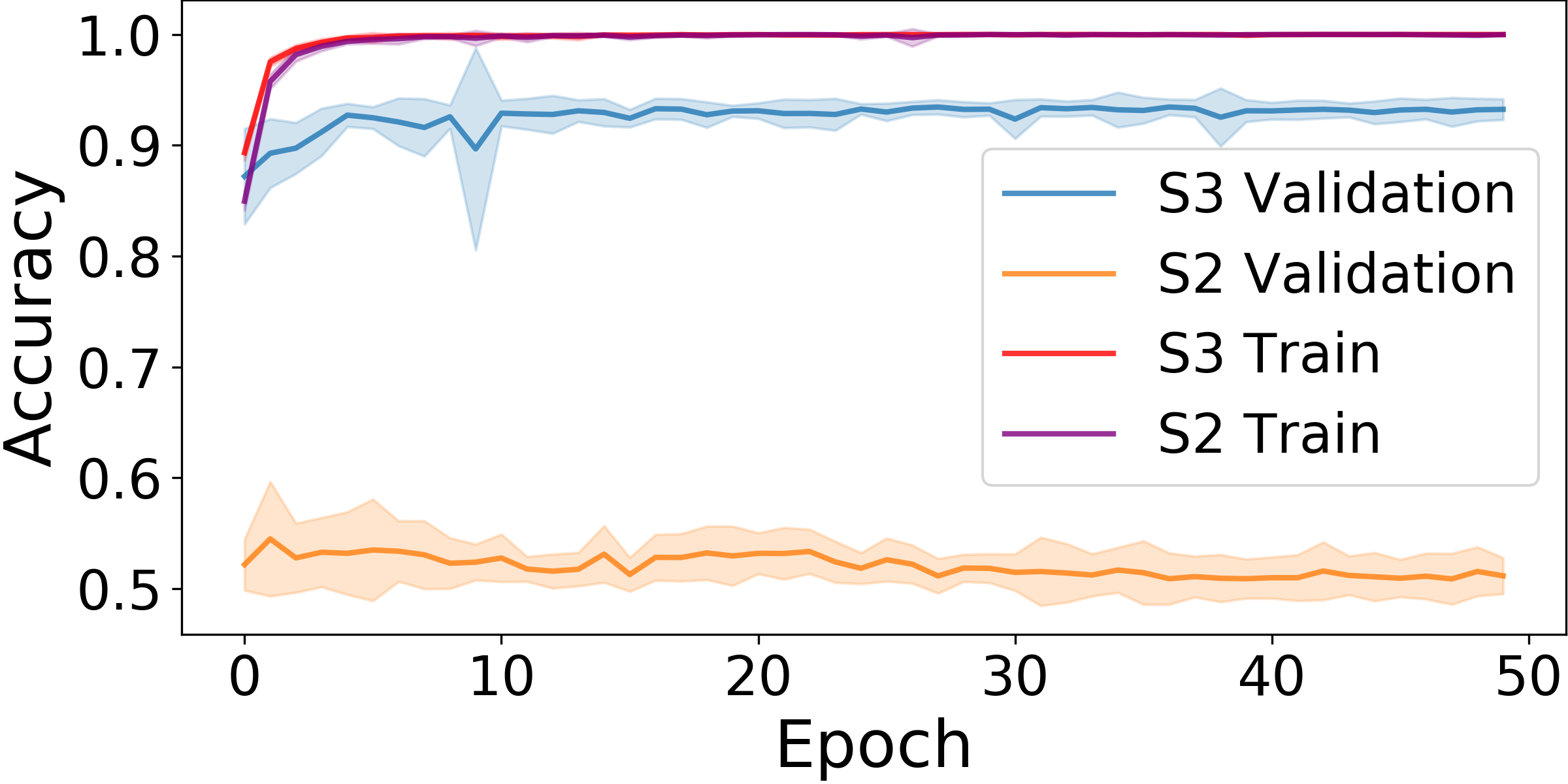}
          \vskip-2mm
          \caption{Diseased}
      \end{subfigure}
      \begin{subfigure}[b]{0.24\textwidth}
          \centering
          \includegraphics[width=1\columnwidth]{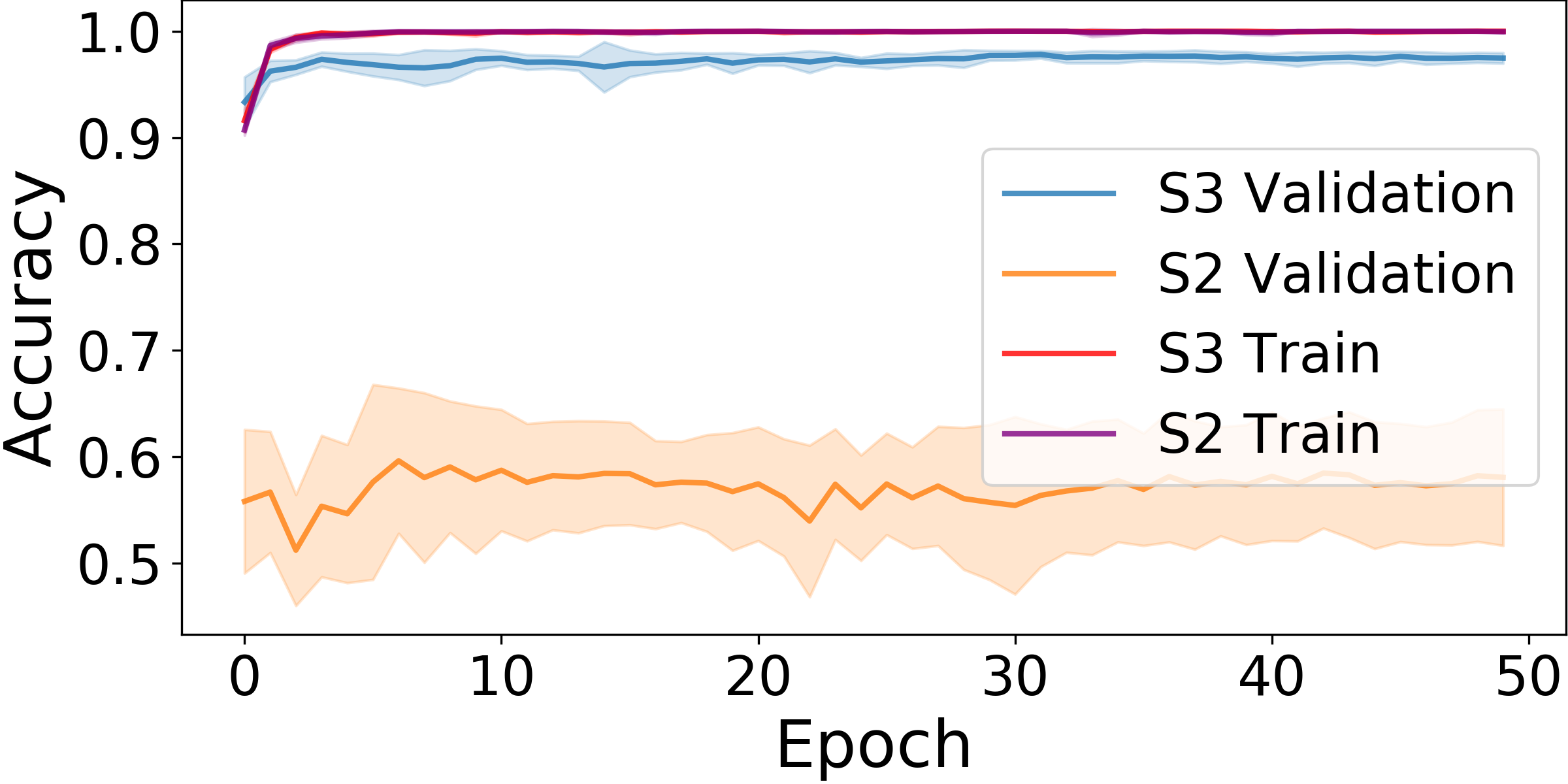}
          \vskip-2mm
          \caption{Contact lens \& printed}
      \end{subfigure}
      \begin{subfigure}[b]{0.24\textwidth}
          \centering
          \includegraphics[width=1\columnwidth]{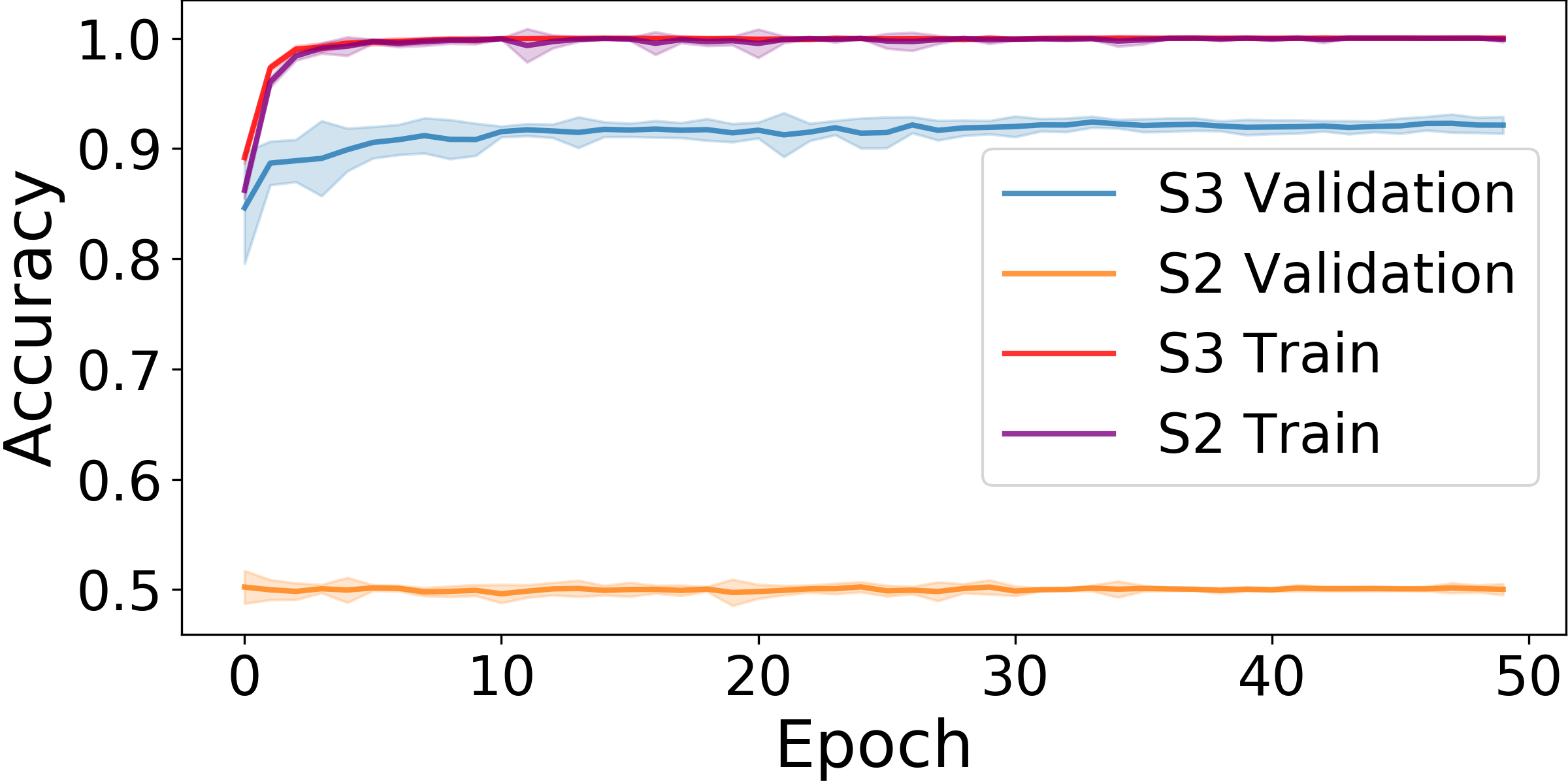}
          \vskip-2mm
          \caption{None (for LivDet testing)}
      \end{subfigure}
  \end{subfigure}
  \caption{Accuracy on the training and validation subsets while training the models without (scenario \textcolor{s2}{\bf S2}) and with (\textcolor{s3}{\bf S3}) human saliency maps encoded into the training data. We can observe a severe overfit, and a close-to-random-chance performance of models trained without employing human intelligence (\textcolor{s2}{\bf S2}). In turn, human-aided saliency maps help in obtaining high validation accuracy almost instantly during training (\textcolor{s3}{\bf S3}).}
  \label{fig:losses}
\end{figure*}

\noindent{\bf Improvement through human-saliency-aided training.}
Since the evaluation 
is focused to assess generalization capabilities,
all samples of one abnormal type are left out in each experiment. Each plot in Fig. \ref{fig:roc} represents one leave-one-out experiment, with the class that was held out indicated in the graph's title. For all experiments, 
each network initiated from pre-trained ImageNet weights 
was trained ten times independently. 
The ROC curves represent the means of ten runs in a given scenario.
The shaded region surrounding each curve represents $\pm 1$ standard deviation of the True Positive Rate (TPR) obtained on the test set.

In all graphs, a similar observation can be made. The {\it large data} models, trained with the largest-possible dataset (scenario \textcolor{s1}{\bf S1}), 
results in the highest accuracy.
That is, domain knowledge about abnormality delivered by dataset of ample size allows for more accurate class representations to be learned.
Models trained with a limited number of samples, and with no human-aided augmentation (scenario \textcolor{s2}{\bf S2}) have lowest accuracy for each left out abnormal type. This clearly demonstrates that such limited training set, although non-trivial to collect if we think about collecting more than 700 exemplars of real attacks on biometric systems, is not sufficient to build an effective PAD system. 

Can we do better with this limited data? The ROC plots show that the use of human-guided features (scenario \textcolor{s3}{\bf S3}) provides a significant increase in accuracy, when compared to scenario \textcolor{s2}{\bf S2}, across all experiments. 
This shows that using human judgement about which parts of the image contain information that is salient to the decision
guides networks to learn solutions that generalize better.
While only trained and tested in the iris PAD context, this approach does not consider any biometrics-specific training or annotations, hence we hypothesise that a similar approach can be applied to a wide family of visual tasks, in which humans present better-than-random-chance classification accuracy. It is important to note that direct comparison of the large data (scenario \textcolor{s1}{\bf S1}) models with human-aided (scenario \textcolor{s3}{\bf S3}) models would not be fair due to significant discrepancies in the training set size.

Varying performance can be seen across the left-out abnormal types. Abnormal types such as textured contacts and synthetic are most difficult due to their intentional resemblance to bona fide samples. When these types are removed from training and validation, the models struggle to make effective classifications on these samples. This is a common observation in iris PAD, as shown in results of the LivDet-Iris 2020 competition. However, in all cases the performance is significantly increased when human annotations are incorporated into training.

\vskip1mm\noindent{\bf Testing on LivDet-Iris 2020 Competition data.} The final evaluation demonstrates how the proposed human-aided training strategy performs on the most recent iris PAD competition. This competition included abnormal types unknown from previously published works, and hence is inherently focused on assessing generalization capabilities of iris PAD algorithms. In this evaluation, we again test models trained according to three scenarios (\textcolor{s1}{\bf S1}, \textcolor{s2}{\bf S2} and \textcolor{s3}{\bf S3}) 
and test on the previously unseen LivDet-Iris 2020 dataset 
using the classification threshold defined by the competition organizers of 0.5, where 0 is bona fide and 1 is abnormal. For this experiment, no abnormal type from the training data was left out. The competition protocol is applied such that we assume no knowledge of the testing set during training. 

As shown in Table \ref{tab:dataset} there are five abnormal image types present in the LivDet dataset, and also bona fide data. One of the abnormal image types (display attack) 
was not present in the training data. The other four abnormal image types were represented in the training with disjoint samples, but were collected by different teams, with different subjects, and were excluded from any training. 
To ensure fairness, no modifications were made to the algorithm to improve results after attaining performance metrics on the LivDet-Iris 2020 benchmark.
All baselines were disqualified from the competition by LivDet organizers since these institutions had access to this test data which originated from the same source as their train data, unlike the competitors. Hence, it was decided to strictly follow the LivDet competition protocol and compare only to the competitors.

As shown in Table \ref{tab:livdet_results}, the highest accuracy is again the ``large data'' model 
(scenario \textcolor{s1}{\bf S1}). This model improves upon the best results in the competition by decreasing the average error rate from the winning level of 29.78\% to a much lower 11.1\%. It is certainly expected, as the large and balanced training set, composed of 93,190 (46,595 bona fide/46,595 abnormal) samples contains the richest information about the task domain. 
However, the most interesting comparison is to juxtapose the ``limited data'' and ``human-aided'' approaches. We can see that using only 765 training images encoded with human saliency maps (scenario \textcolor{s3}{\bf S3}), the average classification error rate (ACER) decreased from 29.78\% (obtained by the competition winner) to 16.37\%. In contrast, when using the same training images but without human-aided training (scenario \textcolor{s2}{\bf S2}) the obtained ACER = 48.37\%, which is worse than the top two LivDet submissions and only marginally better than the last place submission. This suggests that {\bf the human guided approach (\textcolor{s3}{\bf S3}) 
would have won the LivDet-Iris 2020 competition} by a large margin of 13\% whereas an equivalent solution without human annotation incorporation (\textcolor{s2}{\bf S2}) would have placed third.

\vskip2mm\noindent{\bf Validation accuracy increases significantly using models trained with human-aided saliency maps.} 
Figure \ref{fig:losses} outlines an interesting finding when analysing validation accuracies as the training progresses in both \textcolor{s2}{\bf S2} and \textcolor{s3}{\bf S3} scenarios. While the training accuracy in both scenarios stays practically identical, 
it is clear that regular training (\textcolor{s2}{\bf S2}) leads to a severe overfit to the training data, as performance on the validation data is close to random chance and plateaus quickly. 
Conversely, the validation accuracy on the same set when the models are trained with human saliency encoded into the training data (\textcolor{s3}{\bf S3}) is significantly better, and can be achieved rapidly during the training process. It's yet another demonstration that the addition of human saliency information to the training data reduces overfitting, while increasing performance over models trained with the same images without human saliency information encoded.
\section{Summary and Conclusions}

We propose a novel framework for incorporating human saliency judgements into the training images for deep learning, with the goal of increasing accuracy, especially from limited training data, and improving generalization. We validated our approach on the difficult problem of iris presentation attack detection. Iris PAD is an excellent example of a security-critical task that inherently has limited training data available, and for which generalization is of paramount importance. Adversaries create an ever-evolving landscape of attack samples, and algorithms must be able to generalize to novel attacks with only a handful of new images. 

We collected a unique dataset of human annotations from 150 non-expert subjects who highlighted salient regions for this visual classification task. These annotations fed into localized blurring of image regions judged as less salient for humans. The result is a human-saliency-transformed version of the original training images. We experimentally compared the performance of the model trained with (a) human-saliency-transformed data, and (b) original images (with standard augmentation techniques). 
Training with human-saliency-transformed images achieves significantly greater accuracy on all leave-one-attack-type-out experiments. Further, evaluating our approach in the framework of the state-of-the-art LivDet-Iris 2020 competition, our human-saliency-transformed model achieves average classification error rate of 16.37\% on the challenging LivDet-Iris 2020 benchmark, a substantial improvement over the competition winner's ACER = 29.78\%. 

No part of our approach is specific to iris PAD and it can readily be applied to any visual classification task recognizable to humans. This opens a whole new area of research related to effective incorporation of human saliency judgements into training strategies for deep learning. We release our human annotation dataset collected for this work along with source codes to prepare human-aided training data.

{\small
\bibliographystyle{ieee_fullname}
\bibliography{egbib}

\begin{thebibliography}{10}\itemsep=-1pt

\bibitem{casia-database}
Chinese academy of sciences institute of automation.
\newblock http://www.cbsr.ia.ac.cn/china/Iris\%20Databases\%20CH.asp.
\newblock Accessed: 03-12-2021.

\bibitem{Bahdanau_ICLR_2015}
Dzmitry Bahdanau, {Kyung Hyun} Cho, and Yoshua Bengio.
\newblock Neural machine translation by jointly learning to align and
  translate.
\newblock Jan. 2015.
\newblock 3rd International Conference on Learning Representations, ICLR 2015 ;
  Conference date: 07-05-2015 Through 09-05-2015.

\bibitem{Bello_ICCV_2019}
I. {Bello}, B. {Zoph}, Q. {Le}, A. {Vaswani}, and J. {Shlens}.
\newblock Attention augmented convolutional networks.
\newblock In {\em 2019 IEEE/CVF International Conference on Computer Vision
  (ICCV)}, pages 3285--3294, 2019.

\bibitem{Benenson_CVPR_2019}
R. {Benenson}, S. {Popov}, and V. {Ferrari}.
\newblock Large-scale interactive object segmentation with human annotators.
\newblock In {\em 2019 IEEE/CVF Conference on Computer Vision and Pattern
  Recognition (CVPR)}, pages 11692--11701, 2019.

\bibitem{Berga_ELCVIA_2020}
David Berga.
\newblock Understanding eye movements: Psychophysics and a model of primary
  visual cortex.
\newblock {\em {ELCVIA Electronic Letters on Computer Vision and Image
  Analysis}}, 18(2):13--15, 2020.

\bibitem{boyd2019deep}
Aidan Boyd, Adam Czajka, and Kevin Bowyer.
\newblock Deep learning-based feature extraction in iris recognition: Use
  existing models, fine-tune or train from scratch?
\newblock In {\em 2019 IEEE 10th International Conference on Biometrics Theory,
  Applications and Systems (BTAS)}, pages 1--9. IEEE, 2019.

\bibitem{Boyd_PRL_2020}
Aidan Boyd, Zhaoyuan Fang, Adam Czajka, and Kevin~W. Bowyer.
\newblock Iris presentation attack detection: Where are we now?
\newblock {\em Pattern Recognition Letters}, 138:483--489, 2020.

\bibitem{Chen_WACVW_2021}
Cunjian Chen and A. {Ross}.
\newblock An explainable attention-guided iris presentation attack detector.
\newblock In {\em Workshop on Explainable \& Interpretable Artificial
  Intelligence for Biometrics (xAI4Biometrics) at the IEEE Winter Conference on
  Applications of Computer Vision (WACV)}, January 2021.

\bibitem{Cubuk_CVPR_2019}
E.~D. {Cubuk}, B. {Zoph}, D. {Mané}, V. {Vasudevan}, and Q.~V. {Le}.
\newblock Autoaugment: Learning augmentation strategies from data.
\newblock In {\em 2019 IEEE/CVF Conference on Computer Vision and Pattern
  Recognition (CVPR)}, pages 113--123, 2019.

\bibitem{Czajka_WACV_2019}
A. {Czajka}, D. {Moreira}, K. {Bowyer}, and P. {Flynn}.
\newblock Domain-specific human-inspired binarized statistical image features
  for iris recognition.
\newblock In {\em 2019 IEEE Winter Conference on Applications of Computer
  Vision (WACV)}, pages 959--967, 2019.

\bibitem{Das_IJCB_2020}
P. {Das}, J. {Mcfiratht}, Z. {Fang}, A. {Boyd}, G. {Jang}, A. {Mohammadi}, S.
  {Purnapatra}, D. {Yambay}, S. {Marcel}, M. {Trokielewicz}, P. {Maciejewicz},
  K. {Bowyer}, A. {Czajka}, S. {Schuckers}, J. {Tapia}, S. {Gonzalez}, M.
  {Fang}, N. {Damer}, F. {Boutros}, A. {Kuijper}, R. {Sharma}, C. {Chen}, and
  A. {Ross}.
\newblock {Iris Liveness Detection Competition (LivDet-Iris) - The 2020
  Edition}.
\newblock In {\em 2020 IEEE International Joint Conference on Biometrics
  (IJCB)}, pages 1--9, 2020.

\bibitem{Galbally_ICB_2012}
Javier Galbally, Jaime Ortiz-Lopez, Julian Fierrez, and Javier Ortega-Garcia.
\newblock Iris liveness detection based on quality related features.
\newblock In {\em 2012 5th IAPR Int. Conf. on Biometrics (ICB)}, pages
  271--276, New Delhi, India, March 2012. IEEE.

\bibitem{Guo_2019}
Guodong Guo and Na Zhang.
\newblock A survey on deep learning based face recognition.
\newblock {\em Computer Vision and Image Understanding}, 189, 2019.

\bibitem{He_ICCV_2019}
S. {He}, H.~R. {Tavakoli}, A. {Borji}, and N. {Pugeault}.
\newblock Human attention in image captioning: Dataset and analysis.
\newblock In {\em 2019 IEEE/CVF International Conference on Computer Vision
  (ICCV)}, pages 8528--8537, 2019.

\bibitem{Hoffer_CVPR_2020}
E. {Hoffer}, T. {Ben-Nun}, I. {Hubara}, N. {Giladi}, T. {Hoefler}, and D.
  {Soudry}.
\newblock Augment your batch: Improving generalization through instance
  repetition.
\newblock In {\em 2020 IEEE/CVF Conference on Computer Vision and Pattern
  Recognition (CVPR)}, pages 8126--8135, 2020.

\bibitem{Huang_2017_CVPR}
Gao Huang, Zhuang Liu, Laurens van~der Maaten, and Kilian~Q. Weinberger.
\newblock Densely connected convolutional networks.
\newblock In {\em Proceedings of the IEEE Conference on Computer Vision and
  Pattern Recognition (CVPR)}, July 2017.

\bibitem{ISO_19794_6_2011}
{ISO/IEC 19794-6:2011}.
\newblock {Information technology -- Biometric data interchange formats -- Part
  6: Iris image data}, 2011.

\bibitem{Kar_ICCV_2019}
A. {Kar}, A. {Prakash}, M. {Liu}, E. {Cameracci}, J. {Yuan}, M. {Rusiniak}, D.
  {Acuna}, A. {Torralba}, and S. {Fidler}.
\newblock Meta-sim: Learning to generate synthetic datasets.
\newblock In {\em 2019 IEEE/CVF International Conference on Computer Vision
  (ICCV)}, pages 4550--4559, 2019.

\bibitem{Ko_CVPR_2020}
B. {Ko} and G. {Gu}.
\newblock Embedding expansion: Augmentation in embedding space for deep metric
  learning.
\newblock In {\em 2020 IEEE/CVF Conference on Computer Vision and Pattern
  Recognition (CVPR)}, pages 7253--7262, 2020.

\bibitem{Kohli_ICB_2013}
Naman Kohli, Daksha Yadav, Mayank Vatsa, and Richa Singh.
\newblock Revisiting iris recognition with color cosmetic contact lenses.
\newblock In {\em {IEEE} Int. Conf. on Biometrics (ICB)}, pages 1--7, Madrid,
  Spain, June 2013. IEEE.

\bibitem{Kohli_BTAS_2016}
Naman Kohli, Daksha Yadav, Mayank Vatsa, Richa Singh, and Afzel Noore.
\newblock Detecting medley of iris spoofing attacks using desist.
\newblock In {\em {IEEE} Int. Conf. on Biometrics: Theory Applications and
  Systems (BTAS)}, pages 1--6, Niagara Falls, NY, USA, Sept 2016. IEEE.

\bibitem{ETPAD_v2_URL}
Oleg Komogortsev.
\newblock {Eye Tracker Print-Attack Database (ETPAD) v2}, 2014.

\bibitem{Koutilya_CVPR_2020}
P.~N. V.~R. {Koutilya}, H. {Zhou}, and D. {Jacobs}.
\newblock Sharingan: Combining synthetic and real data for unsupervised
  geometry estimation.
\newblock In {\em 2020 IEEE/CVF Conference on Computer Vision and Pattern
  Recognition (CVPR)}, pages 13971--13980, 2020.

\bibitem{Sung_OE_2007}
Sung~Joo Lee, Kang~Ryoung Park, Youn~Joo Lee, Kwanghyuk Bae, and Jai~Hie Kim.
\newblock {Multifeature-based fake iris detection method}.
\newblock {\em Optical Engineering}, 46(12):1 -- 10, 2007.

\bibitem{Li_JV_2007}
Fei-Fei Li, Asha Iyer, Christof Koch, and Pietro Perona.
\newblock What do we perceive in a glance of a real-world scene?
\newblock {\em Journal of Vision}, 7(10), 2007.

\bibitem{Litjens_2017}
Geert Litjens, Thijs Kooi, Babak~Ehteshami Bejnordi, Arnaud Arindra, Adiyoso
  Setio, Francesco Ciompi, Mohsen Ghafoorian, Jeroen~A.W.M. van~der Laak, Bram
  van Ginneken, and Clara~I. Sánchez.
\newblock A survey on deep learning in medical image analysis.
\newblock {\em Medical Image Analysis}, 42:60—88, 2017.

\bibitem{Liu_IJCV_2020}
Li Liu, Wanli Ouyang, Xiaogang Wang, Paul Fieguth, Jie Chen, Xinwang Liu, and
  Matti Pietik{\"a}inen.
\newblock Deep learning for generic object detection: A survey.
\newblock {\em International Journal of Computer Vision}, 128(2):261--318, Feb
  2020.

\bibitem{Luo_CVPR_2020}
C. {Luo}, Y. {Zhu}, L. {Jin}, and Y. {Wang}.
\newblock Learn to augment: Joint data augmentation and network optimization
  for text recognition.
\newblock In {\em 2020 IEEE/CVF Conference on Computer Vision and Pattern
  Recognition (CVPR)}, pages 13743--13752, 2020.

\bibitem{Masi_2018}
I. {Masi}, Y. {Wu}, T. {Hassner}, and P. {Natarajan}.
\newblock Deep face recognition: A survey.
\newblock In {\em 2018 31st SIBGRAPI Conference on Graphics, Patterns and
  Images (SIBGRAPI)}, pages 471--478, 2018.

\bibitem{Moreira_WACV_2019}
D. {Moreira}, M. {Trokielewicz}, A. {Czajka}, K. {Bowyer}, and P. {Flynn}.
\newblock Performance of humans in iris recognition: The impact of iris
  condition and annotation-driven verification.
\newblock In {\em 2019 IEEE Winter Conference on Applications of Computer
  Vision (WACV)}, pages 941--949, 2019.

\bibitem{OToole_TSMC_2007}
A.~J. {O'Toole}, H. {Abdi}, F. {Jiang}, and P.~J. {Phillips}.
\newblock Fusing face-verification algorithms and humans.
\newblock {\em IEEE Transactions on Systems, Man, and Cybernetics, Part B
  (Cybernetics)}, 37(5):1149--1155, 2007.

\bibitem{Peterson_ICCV_2019}
J. {Peterson}, R. {Battleday}, T. {Griffiths}, and O. {Russakovsky}.
\newblock Human uncertainty makes classification more robust.
\newblock In {\em 2019 IEEE/CVF International Conference on Computer Vision
  (ICCV)}, pages 9616--9625, 2019.

\bibitem{RichardWebster_TPAMI_2018}
B. {Richard Webster}, S.~E. {Anthony}, and W.~J. {Scheirer}.
\newblock Psyphy: A psychophysics driven evaluation framework for visual
  recognition.
\newblock {\em IEEE Transactions on Pattern Analysis and Machine Intelligence},
  41(9):2280--2286, 2019.

\bibitem{RichardWebster_ECCV_2018}
Brandon RichardWebster, So~Yon Kwon, Christopher Clarizio, Samuel~E. Anthony,
  and W. Scheirer.
\newblock Visual psychophysics for making face recognition algorithms more
  explainable.
\newblock In {\em European Conference on Computer Vision (ECCV)}, pages
  263--281, 2018.

\bibitem{Sharma_IJCB_2020}
R. {Sharma} and A. {Ross}.
\newblock {D-NetPAD: An Explainable and Interpretable Iris Presentation Attack
  Detector}.
\newblock In {\em 2020 IEEE International Joint Conference on Biometrics
  (IJCB)}, pages 1--10, 2020.

\bibitem{Shorten_JBD_2019}
Connor Shorten and Taghi~M. Khoshgoftaar.
\newblock A survey on image data augmentation for deep learning.
\newblock {\em Journal of Big Data}, 6(1):60, Jul 2019.

\bibitem{Sundararajan_CSUR_2018}
Kalaivani Sundararajan and Damon~L. Woodard.
\newblock Deep learning for biometrics: A survey.
\newblock {\em ACM Comput. Surv.}, 51(3), May 2018.

\bibitem{Trokielewicz_BTAS_2015}
M. {Trokielewicz}, A. {Czajka}, and P. {Maciejewicz}.
\newblock Assessment of iris recognition reliability for eyes affected by
  ocular pathologies.
\newblock In {\em {IEEE} Int. Conf. on Biometrics: Theory Applications and
  Systems (BTAS)}, pages 1--6, 2015.

\bibitem{Trokielewicz_BTAS_2019}
M. {Trokielewicz}, A. {Czajka}, and P. {Maciejewicz}.
\newblock Perception of image features in post-mortem iris recognition: Humans
  vs machines.
\newblock In {\em 2019 IEEE 10th International Conference on Biometrics Theory,
  Applications and Systems (BTAS)}, pages 1--8, 2019.

\bibitem{Trokielewicz_IVC_2020}
Mateusz Trokielewicz, Adam Czajka, and Piotr Maciejewicz.
\newblock Post-mortem iris recognition with deep-learning-based image
  segmentation.
\newblock {\em Image and Vision Computing}, 94:103866, 2020.

\bibitem{Tsirikoglou_CGF_2020}
A. Tsirikoglou, G. Eilertsen, and J. Unger.
\newblock A survey of image synthesis methods for visual machine learning.
\newblock {\em Computer Graphics Forum}, 39(6):426--451, 2020.

\bibitem{Wang_CVPR_2019}
Q. {Wang}, J. {Gao}, W. {Lin}, and Y. {Yuan}.
\newblock Learning from synthetic data for crowd counting in the wild.
\newblock In {\em 2019 IEEE/CVF Conference on Computer Vision and Pattern
  Recognition (CVPR)}, pages 8190--8199, 2019.

\bibitem{WARSAW_DBs_URL}
{Warsaw University of Technology}.
\newblock {Warsaw Datasets Webpage}.
\newblock http://zbum.ia.pw.edu.pl/EN/node/46, 2013.

\bibitem{Wei_ICPR_2008}
Zhuoshi Wei, Tieniu Tan, and Zhenan Sun.
\newblock Synthesis of large realistic iris databases using patch-based
  sampling.
\newblock In {\em Int. Conf. on Pattern Recognition (ICPR)}, pages 1--4, Tampa,
  FL, USA, Dec 2008. IEEE.

\bibitem{Yambay_IJCB_2017}
David Yambay, Benedict Becker, Naman Kohli, Daksha Yadav, Adam Czajka, Kevin~W.
  Bowyer, Stephanie Schuckers, Richa Singh, Mayank Vatsa, Afzel Noore, Diego
  Gragnaniello, C. Sansone, L. Verdoliva, Lingxiao He, Yiwei Ru, Haiqing Li,
  Nianfeng Liu, Zhenan Sun, and Tieniu Tan.
\newblock {LivDet Iris 2017 -- Iris Liveness Detection Competition 2017}.
\newblock In {\em {IEEE} Int. Joint Conf. on Biometrics (IJCB)}, pages 1--6,
  Denver, CO, USA, 2017. IEEE.

\bibitem{Yambay_ISBA_2017}
David Yambay, Brian Walczak, Stephanie Schuckers, and Adam Czajka.
\newblock {LivDet-Iris 2015 - Iris Liveness Detection Competition 2015}.
\newblock In {\em {IEEE} Int. Conf. on Identity, Security and Behavior Analysis
  (ISBA)}, pages 1--6, New Delhi, India, Feb 2017. IEEE.

\end{thebibliography}
}
\appendix

\section{Detailed Dataset Description}
\vspace{-8em}
\begin{table}[]
\caption{Full dataset used for \textbf{training and validation} broken down by individual contributing dataset. The in-house currently unpublished data used in this work is denoted as \textit{University of Notre Dame data}.}
\centering
\label{app:full_dataset}
\begin{tabular}{cccc}
\hline
\textbf{Image Type}                    & \textbf{Contributing Dataset}                                                                                                                                                                                                                                                                                                                    & \textbf{\# of Samples}                                                                                                                     & \textbf{Total Samples} \\ \hline
\textbf{Bona fide}                          & \begin{tabular}[c]{@{}c@{}}
ATVS-FIr \cite{Galbally_ICB_2012}\\ 
BERC\_IRIS\_FAKE \cite{Sung_OE_2007}\\ 
CASIA-Iris-Thousand \cite{casia-database}\\ 
CASIA-Iris-Twins \cite{casia-database}\\ 
Disease-Iris v2.1 \cite{Trokielewicz_BTAS_2015}\\ 
ETPAD v2 {\cite{ETPAD_v2_URL}}\\ 
IIITD Contact Lens Iris \cite{Kohli_ICB_2013}\\ 
IIITD Combined Spoofing Database \cite{Kohli_BTAS_2016} \\ 
LivDet-Iris Clarkson 2015 \cite{Yambay_ISBA_2017} \\ 
LivDet-Iris Warsaw 2015 \cite{Yambay_ISBA_2017}\\ 
LivDet-Iris Clarkson 2017 \cite{Yambay_IJCB_2017} \\ 
LivDet-Iris IIITD-WVU 2017 \cite{Yambay_IJCB_2017}\\ 
LivDet-Iris Warsaw 2017 \cite{Yambay_IJCB_2017}\\ 
University of Notre Dame data\end{tabular} 
& \begin{tabular}[c]{@{}c@{}}800\\ 2,776\\ 19,952\\ 3,181\\ 255\\ 400\\ 13\\ 4,531\\ 813\\ 36\\ 3,949\\ 2,944\\ 5,167\\ 354,236\end{tabular} & \textbf{399,053}       \\ \hline
\textbf{Textured contact lens}         & \begin{tabular}[c]{@{}c@{}}
BERC\_IRIS\_FAKE \cite{Sung_OE_2007}\\ 
IIITD Contact Lens Iris \cite{Kohli_ICB_2013}\\ 
LivDet-Iris Clarkson 2015 \cite{Yambay_ISBA_2017} \\ 
LivDet-Iris Clarkson 2017 \cite{Yambay_IJCB_2017}\\ 
LivDet-Iris IIITD-WVU 2017 \cite{Yambay_IJCB_2017}\\ 
University of Notre Dame data\end{tabular}                                                                                                                                                    & \begin{tabular}[c]{@{}c@{}}140\\ 3,420\\ 1,107\\ 1,881\\ 1,700\\ 19,124\end{tabular}                                                       & \textbf{27,372}        \\ \hline
\textbf{Paper printouts}               & \begin{tabular}[c]{@{}c@{}}
ATVS-FIr \cite{Galbally_ICB_2012}\\ 
BERC\_IRIS\_FAKE \cite{Sung_OE_2007}\\ 
IIITD Combined Spoofing Database \cite{Kohli_BTAS_2016}\\ 
LivDet-Iris Clarkson 2015 \cite{Yambay_ISBA_2017}\\ 
LivDet-Iris Warsaw 2015 \cite{Yambay_ISBA_2017}\\ 
LivDet-Iris Clarkson 2017 \cite{Yambay_IJCB_2017}\\ 
LivDet-Iris IIITD-WVU 2017 \cite{Yambay_IJCB_2017}\\ 
LivDet-Iris Warsaw 2017 \cite{Yambay_IJCB_2017}\end{tabular}                                                                                                       & \begin{tabular}[c]{@{}c@{}}800\\ 1,600\\ 1,371\\ 1,745\\ 20\\ 2,250\\ 1,766\\ 6,841\end{tabular}                                           & \textbf{16,393}        \\ \hline
\textbf{Post-mortem Irises}            & Post-Mortem-Iris v3.0 \cite{Trokielewicz_IVC_2020}                                                                                                                                                                                                                                                                                                                           & 2,259                                                                                                                                      & \textbf{2,259}         \\ \hline
\textbf{Synthetic}                     & CASIA-Iris-Syn V4 \cite{Wei_ICPR_2008}                                                                                                                                                                                                                                                                                                                                 & 10,000                                                                                                                                     & \textbf{10,000}        \\ \hline
\textbf{Artificial}                    & \begin{tabular}[c]{@{}c@{}}BERC\_IRIS\_FAKE \cite{Sung_OE_2007} \\ University of Notre Dame data \end{tabular}                                                                                                                                                                                                                                                                   & \begin{tabular}[c]{@{}c@{}}80\\ 197\end{tabular}                                                                                           & \textbf{277}           \\ \hline
\textbf{Diseased irises}               & Disease-Iris v2.1 \cite{Trokielewicz_BTAS_2015}                                                                                                                                                                                                                                                                                                                                    & 1,537                                                                                                                                      & \textbf{1,537}         \\ \hline
\textbf{Textured contacts \& printed} & LivDet-Iris IIITD-WVU 2017 \cite{Yambay_IJCB_2017}                                                                                                                                                                                                                                                                                                         & 1,899                                                                                                                                      & \textbf{1,899}         \\ \hline
\end{tabular}
\end{table}

\newpage

\section{Annotation Tool}
\label{fig:annot_tool}

\begin{figure*}[h!]
    \centering
    \includegraphics[width=1.0\linewidth]{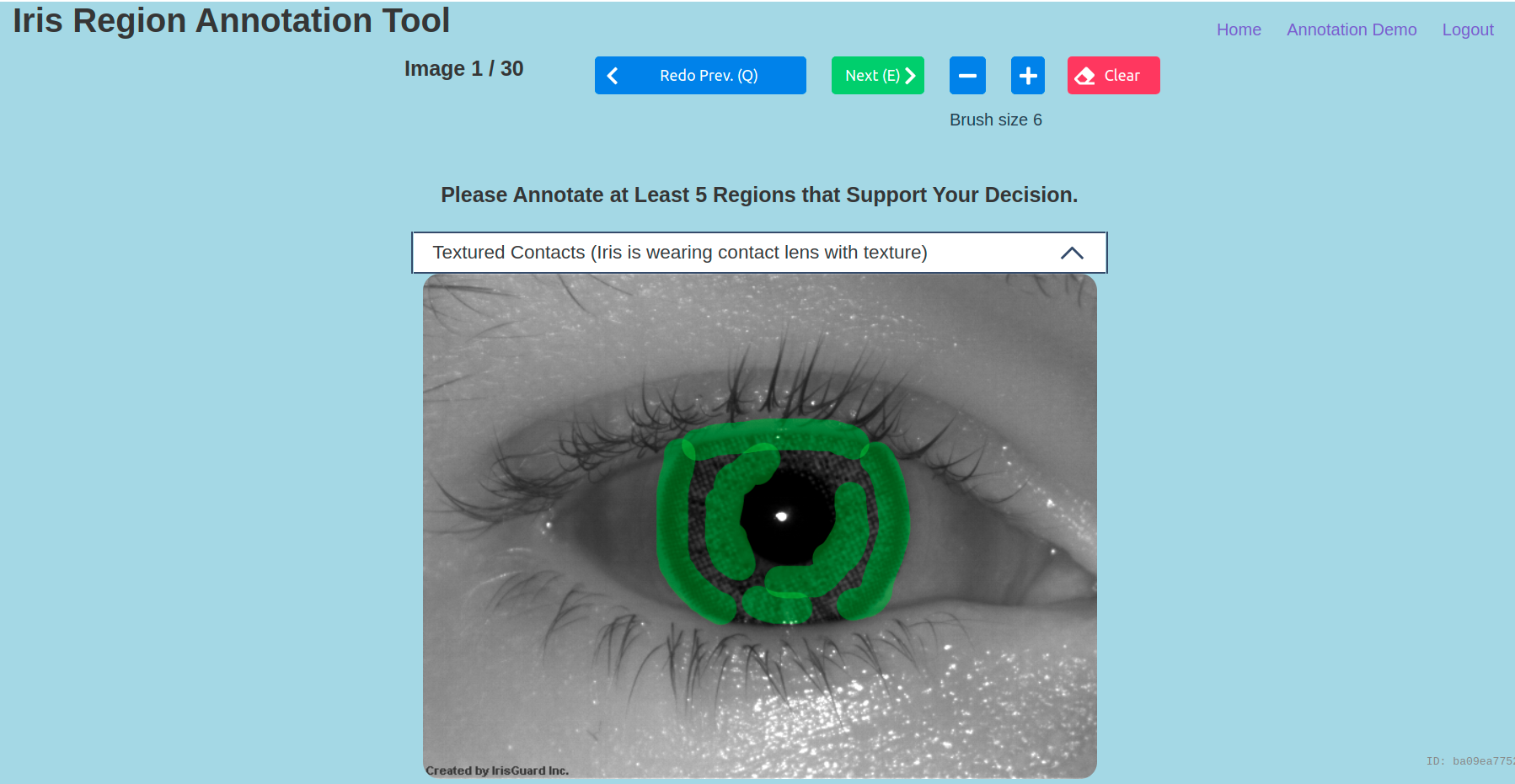}
    \caption{Online annotation tool developed to collect annotation data with an example input of a human solving the iris presentation attack detection task for a textured contact lens sample.}
    \label{fig:gui}
\end{figure*}

\end{document}